%% file: 0.main.tex
\documentclass[11pt]{article}

\usepackage[final]{acl}

\usepackage{times}
\usepackage{latexsym}

\usepackage[T1]{fontenc}

\usepackage[utf8]{inputenc}

\usepackage{microtype}

\usepackage{inconsolata}

\usepackage{graphicx}
\usepackage{graphicx}
\usepackage{multirow} 
\usepackage{booktabs} 
\usepackage{enumitem}
\usepackage{pifont}
\usepackage{float}
\usepackage{ulem}
\usepackage{subcaption}
\usepackage{mwe}
\usepackage{cuted}
\usepackage{caption}
\usepackage{amsmath}
\usepackage{amsfonts}
\usepackage{xspace}
\newcommand{\ourmethod}{{\fontfamily{lmtt}\selectfont \textbf{\textit{DoCtOR}}}\xspace}

\newcommand{\afamethod}{{\fontfamily{lmtt}\selectfont \textbf{\textit{ProFA}}}\xspace}
\newcommand{\Rmnum}[1]{\uppercase\expandafter{\romannumeral #1}}

\title{Finding Where the Buck Stops: An Automated Failure Attribution-Based Reflection Framework for Multi-Agent Collaboration}

\author{
\bf Xiaoqing Wang$^{1}$ \quad
\bf Keman Huang$^{1}$\thanks{~~Keman Huang (keman@ruc.edu.cn) is the corresponding author.} \quad
\bf Bin Liang$^{1}$ \quad
\bf Hongyu Li$^{2}$ \\
\bf Xiaoyong Du$^{1}$ \quad
\bf Wuqiong Pan$^{2}$ \\
$^1$Renmin University of China \\
$^2$Ant Group \\[4pt]
\normalfont
\{wangxiaoq, keman, liangb, duyong\}@ruc.edu.cn
}

\begin{document}
\maketitle
\begin{abstract}
Multi-agent systems (MAS) powered by large language models have shown promise for complex tasks but suffer from high failure rates. Current self-reflection methods for MAS require all agents to reflect upon failure, overlooking a critical reality: failures typically stem from a specific agent leading the task astray, namely the \textit{decisive error agent}, while others merely fulfill their regular duties. Forcing regular-behaving agents to reflect contaminates their memory with wrong insights. 
Hence, we propose \ourmethod (\textbf{D}iagn\textbf{o}se-then-\textbf{C}orrec\textbf{t} PP\textbf{O}-enhanced \textbf{R}eflection), a novel reflection framework that enhances multi-agent collaboration. \ourmethod first identifies the \textit{decisive error step} and \textit{decisive error agent} through automated failure attribution, then employs counterfactual reasoning to generate a \textit{corrected decisive error step}, and finally engages only the \textit{decisive error agent} to produce targeted reflections.
Experimental results show \ourmethod achieves 22\%, 26\%, and 27\% improvements over initial success rates on HotPotQA, ChartQAPro, and Mind2Web datasets, outperforming Reflexion, Retroformer, and COPPER. We further establish the generalizability of our diagnose-then-correct paradigm and demonstrate that in low-resource settings, focusing reflection on reasoning steps after the \textit{decisive error step} achieves comparable quality to reflecting on the complete failure trajectory.
\end{abstract}

\input{1.Introduction}

\input{2.Preliminary}

\input{3.Method}

\input{4.Experiment}

\input{6.Conclusion}

\normalem
\bibliography{custom}

\appendix

\input{7.Appendix}

\end{document}

%% file: 1.Introduction.tex
\section{Introduction}

In recent years, multi-agent systems (MAS) powered by large language models have attracted considerable attention and are increasingly regarded as a promising approach for tackling complex tasks \cite{wu2024autogen,qian2024chatdev,chen2023agentverse,hong2024metagpt,fourney2024magentic}. Intuitively, multiple specialized agents working collaboratively should leverage collective intelligence to overcome challenges insurmountable for single agents.
However, recent empirical studies \cite{wang2024rethinking,zhang2025if,pan2025why} have revealed a perplexing phenomenon: \textit{despite consuming additional computational resources, MAS frequently fail to reliably outperform robust single-agent baselines}. \citet{pan2025why} further documents that prominent MAS, including Chatdev \cite{qian2024chatdev}, MetaGPT \cite{hong2024metagpt}, Magentic-One \cite{fourney2024magentic}, and AppWorld \cite{trivedi2024appworld}, demonstrate failure rates between 60\% and 86.7\%, even on tasks they were specifically designed to excel at. Such alarming failure rates raise fundamental concerns regarding the practical application of MAS.

\begin{figure}[t]
    \centering
    \includegraphics[width=\linewidth]{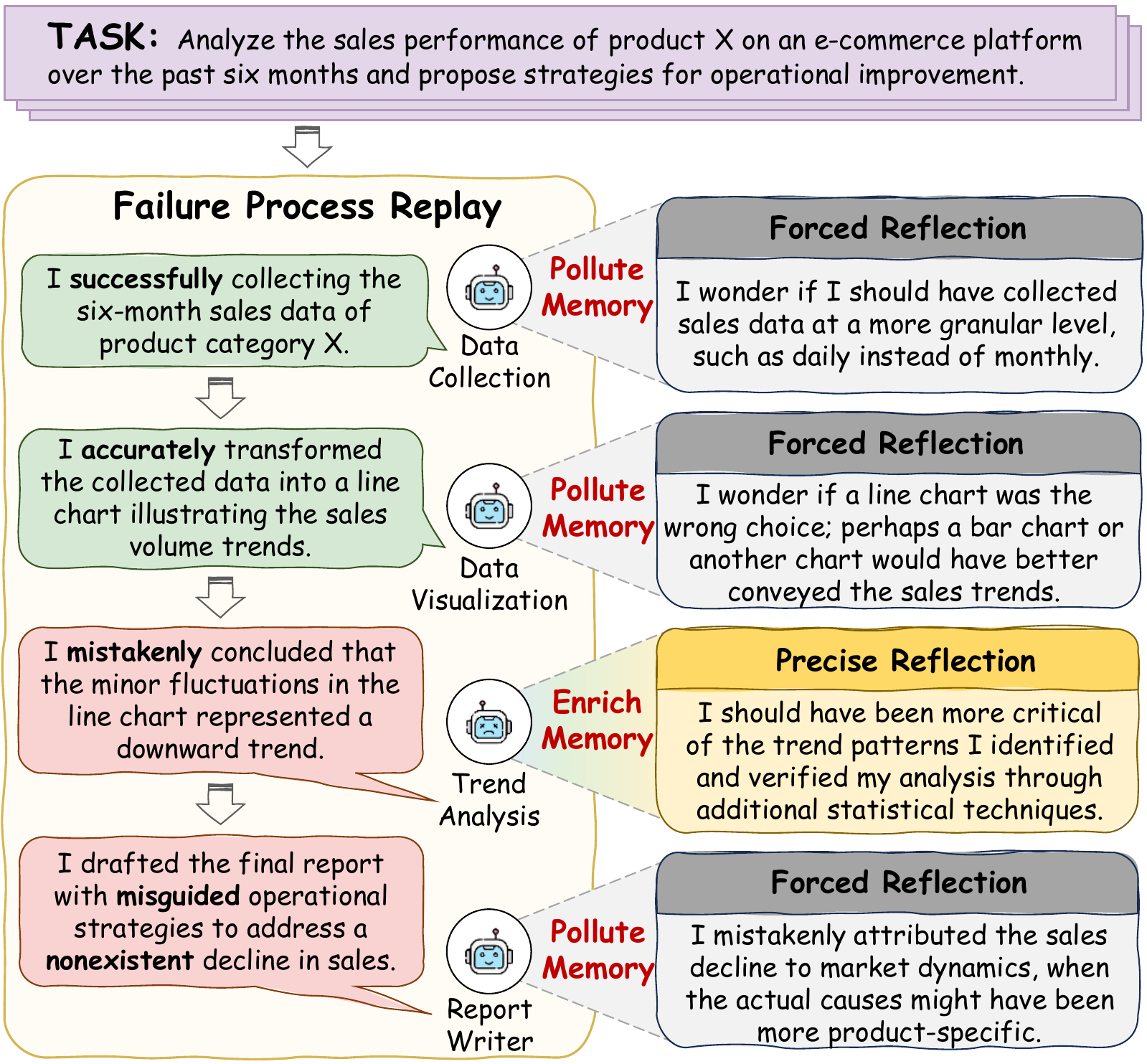}
    \caption{A multi-agent collaboration failure process replay in a data analysis task and the detrimental effects of forced reflection on regular-behaving agents. }
    \label{fig:intro}
\end{figure}

To enhance collaborative performance in MAS, a natural approach involves collecting extensive collaboration data for agent fine-tuning. Nevertheless, this strategy may compromise the model's general capabilities \cite{yang2024unveiling}, which contradicts the goal of achieving artificial general intelligence. Therefore, a more prevalent approach is to optimize the collaboration process through \textbf{\textit{self-reflection}} mechanisms \cite{shinn2023reflexion,yao2024retroformer,bo2024reflective}, which transform binary or scalar rewards from the environment into verbal reflections, thereby providing additional context to improve task performance.

Existing literature has developed self-reflection methods for MAS, such as COPPER \cite{bo2024reflective}, which fine-tunes a shared reflector to generate personalized reflections according to agent roles. Specifically, COPPER requires all participating agents to reflect on task trajectories when failures occur. The underlying assumption of COPPER is that every agent shares responsibility for the failure and should reflect on its causes. However, it overlooks an important reality: \textit{When tasks fail, responsibility often falls on a specific agent leading the task astray, namely the \textit{decisive error agent}, while others merely fulfill their normal duties}.

Figure \ref{fig:intro} illustrates this phenomenon through a multi-agent collaboration failure process replay:
\textit{Agent$_A$} successfully collected the monthly sales data, while \textit{Agent$_B$} correctly generated a corresponding line graph. However, \textit{Agent$_C$} erroneously interpreted minor fluctuations as a downward trend, leading \textit{Agent$_D$} to develop inappropriate strategies based on this misanalysis. 
In this scenario, \textit{Agent$_C$} is the \textit{decisive error agent}, while others are \textit{regular-behaving agents}. \textit{\textbf{Forced reflection can be detrimental for regular-behaving agents}}: \textit{Agent$_A$} might question whether daily data would have been more appropriate, while \textit{Agent$_B$} might wonder if a bar chart would have been superior to the line graph. 
Such misguided reflections contaminate the memory of regular-behaving agents, providing erroneous insights for future task execution and thus probably introducing new errors, such as \textit{Agent$_A$} collecting redundant data or \textit{Agent$_B$} abandoning line chart visualization methods.

Therefore, we propose \textbf{\uwave{D}iagn\uwave{o}se-then-\uwave{C}orrec\uwave{t} PP\uwave{O}-enhanced \uwave{R}eflection (\ourmethod)}, a novel framework for enhancing multi-agent collaboration. The framework consists of an action module for generating reasoning trajectories and a reflection module that provides feedback and updates agent memory.
Within the reflection module, we introduce a lightweight \ding{168} \textbf{diagnosis submodule} for \textit{\textbf{automatic failure attribution}} \cite{zhang2025which}, which identifies the \textit{decisive error step} and the corresponding \textit{decisive error agent} in multi-agent reasoning trajectories. Inspired by Process Reward Models (PRM) \cite{lightman2023let,zhang-etal-2025-lessons,song-etal-2025-prmbench}, we propose a \textbf{\uwave{Pro}cess reward-based Automated \uwave{F}ailure \uwave{A}ttribution (\textit{\afamethod})} method that assigns correctness scores to individual reasoning steps, where the first incorrect step is designated as the decisive error step and its agent as the decisive error agent.
Based on the identified failure points, the \ding{171} \textbf{correction submodule} applies counterfactual reasoning to generate a corrected decisive error step, whose correctness is subsequently verified by \textit{\afamethod}. Finally, only the decisive error agent performs targeted reflection through the \ding{170} \textbf{reflector model}, enabling \textit{self-correction} and \textit{self-improvement}. 
The reflector is further fine-tuned through proximal policy optimization \cite{schulman2017proximal}.

Our contributions can be concluded as follows:
\vspace{-0.6em}
\begin{itemize}[leftmargin=*,itemsep=-0.3em]
\item[\ding{182}] \textbf{\textit{Framework Proposal.}} 
We propose \textbf{\textit{\ourmethod}}, a novel reflection framework that addresses limitations of existing multi-agent reflection methods. Unlike approaches that require all agents to reflect upon failure, \ourmethod first identifies the \textit{decisive error step} through automated failure attribution, then employs counterfactual reasoning to generate the \textit{corrected decisive error step}, and finally engages the \textit{decisive error agent} to generate precise reflections while avoiding memory contamination of regular-behaving agents.

\item[\ding{183}] \textbf{\textit{Experimental Validation.}} 
Experiments show that \textit{\textbf{(\Rmnum{1}) \ourmethod}} exhibits superior reflection capabilities, improving initial success rates by 22\%, 26\%, and 27\% on HotPotQA, ChartQAPro, and Mind2Web datasets respectively, outperforming  baselines including Reflexion, Retroformer, and COPPER; 
\textit{\textbf{(\Rmnum{2}) \afamethod}} provides effective automated failure attribution, achieving 4\%-35\% improvement in agent-level accuracy and 9\%-28\% improvement in step-level accuracy on the Who \& When dataset compared to existing methods.

\item[\ding{184}] \textbf{\textit{Practical Solution.}} 
We demonstrate the generalizability of the \textit{diagnose-then-correct} paradigm in enhancing existing prompt-based reflection methods such as Reflexion.
Additionally, our analysis reveals that providing only the reasoning steps following the \textit{decisive error step} achieves comparable reflection quality to complete trajectories, enabling efficient reflection generation under low-resource settings.

\end{itemize}

%% file: 2.Preliminary.tex
\begin{figure*}[t]
    \centering
    \includegraphics[width=\linewidth]{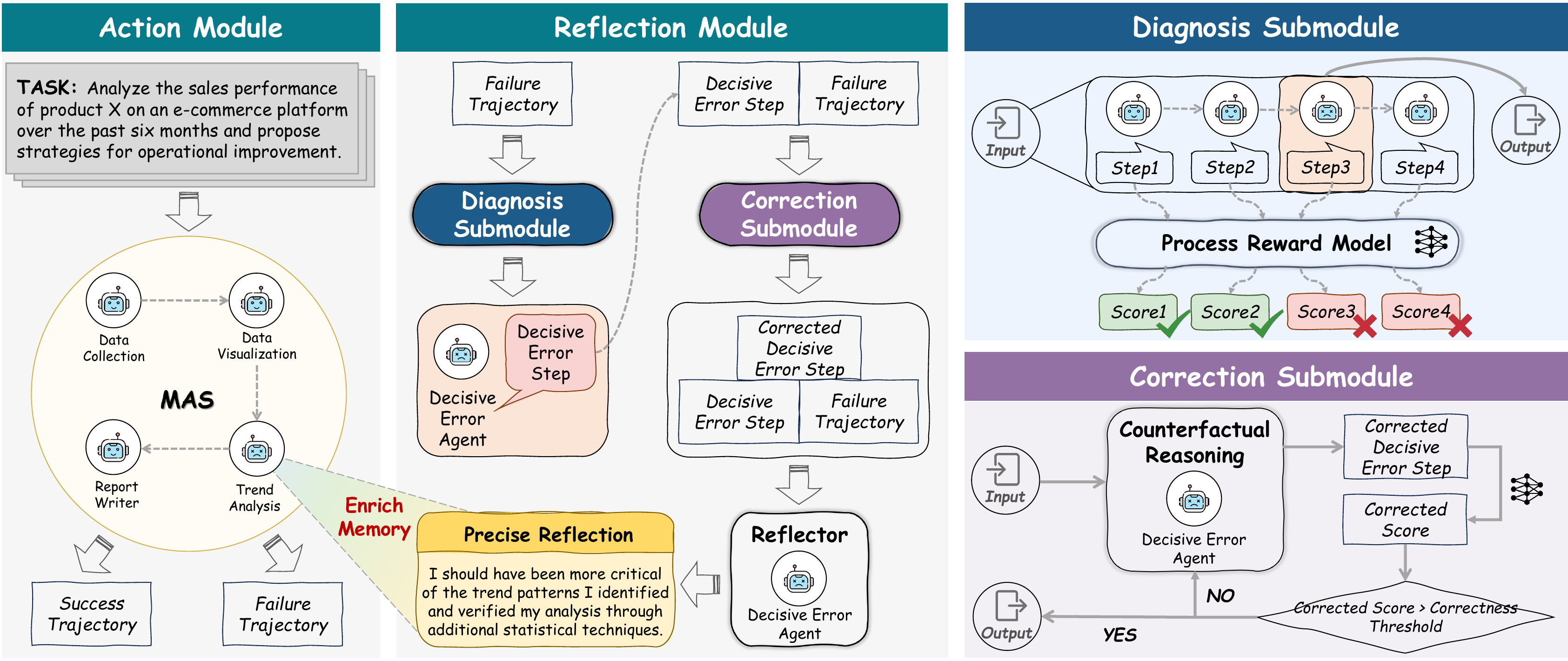}
    \caption{Overview of \textit{\ourmethod} framework. Our framework comprises two components: an action module for generating reasoning trajectories, and a reflection module that provides feedback to action module agents and enriches their memory. The reflection module follows a three-stage workflow: (1) diagnosis submodule identifies the decisive error step and agent; (2) correction submodule generates a corrected decisive error step; and (3) reflector model leverages the previous two stages' outputs to produce targeted reflections for the decisive error agent.}
    
    \label{fig:method}
\end{figure*}

\section{Preliminary}
\subsection{Multi-Agent Collaboration}

In this work, we formalize the LLM-based multi-agent system using a tuple $(N, S, A, P_{\xi_o}, R)$, where $N$ represents the number of agents, $S = S_1 \times S_2 \times \cdots \times S_N$ denotes the joint state space, and $A = A_1 \times A_2 \times \cdots \times A_N$ represents the joint action space. 
Both states $S$ and actions $A$ are described in natural language. At each timestep, the current state $s$ is integrated into a context $c$, which may include descriptions of previous states to provide historical information, enabling LLMs to make more informed decisions about the next action to take. The state transition function $P_{\xi_o}: S \times A \rightarrow S$ governs the system dynamics, where $\xi_o$ captures the inherent randomness in state transitions. 

The multi-agent system executes tasks through sequential interactions with the environment, generating trajectories of the form $\tau = \{s_0, a_0, s_1, a_1, \ldots, s_T, a_T\}$, where $T$ denotes the trajectory length. 
In the cooperative setting, all agents share a common objective and work together to maximize collective performance. Following the standard reinforcement learning paradigm, the ultimate goal is to maximize the cumulative reward, or episode return: $G(\tau) = \sum_{t=0}^{T} R(s_t, a_t)$.

\subsection{Automated Failure Attribution}

To better address failures in multi-agent collaboration, we formulate the problem of automated failure attribution and introduce the concept of \textit{decisive error step} and \textit{decisive error agent}.

For a failed trajectory $\tau$, we define the \textit{decisive error step} $t^*$ as the first time step where an agent's action $a_{t^*}$ is identified as incorrect. This formulation is based on the observation that in multi-agent collaboration, errors often propagate sequentially. When one agent makes a mistake, subsequent agents may build upon this incorrect foundation, leading to a chain of compounding errors. Therefore, identifying and correcting the first error is more effective than attempting to fix multiple downstream mistakes that stem from the same root cause. Correspondingly, the \textit{decisive error agent} $i^*$ is the agent responsible for the erroneous action $a_{t^*}$ at the \textit{decisive error step} $t^*$.

%% file: 3.Method.tex
\section{Method}

As illustrated in Figure \ref{fig:method}, our \textit{\ourmethod} framework consists of an action module and a reflection module. While the action module comprises multiple agents based on frozen language models (GPT-4o-mini) with inaccessible parameters, the reflection module centers on a \textit{reflector model}, a smaller local language model (Llama-3.1-8B-Instruct) that can be fine-tuned in low-resource environments. Detailed prompts for both the action and reflection modules are provided in Appendix \ref{sec:DoCTORprompts}.

In the remainder of this section, we describe the \textbf{reflection module} in detail. Specifally, the \textit{reflector model} ($\triangleright$ Section \ref{sec:reflection}) leverages the decisive error step identified by the \textit{diagnosis submodule} ($\triangleright$ Section \ref{sec:Diagnosis}) and the corrected decisive error step generated by the \textit{correction submodule} ($\triangleright$ Section \ref{sec:correction}), and the failed trajectory to generate targeted reflections for the decisive error agent.

\subsection{Diagnosis Submodule}
\label{sec:Diagnosis}
Inspired by the process reward model  \cite{lightman2023let,zhang-etal-2025-lessons,song-etal-2025-prmbench}, we develop a \afamethod method, which assigns correctness scores to each reasoning step in a multi-agent collaboration trajectory.

Given a failed  trajectory $\tau$ generated by multi agents, \textit{\afamethod} assigns a correctness score $c_t \in [0, 1]$ to each step $t$ in the reasoning process:
\begin{equation}c_t = \afamethod_{\phi}(s_t, a_t,\tau_{<t})\end{equation}

where $\afamethod_{\phi}$ is a process reward model parameterized by $\phi$, and $\tau_{<t}$ represents the trajectory up to step $t$. 
Based on these scores, we formulate a binary correctness indicator $B_t$ for each step:
\begin{equation}B_t = \mathbb{I}(c_t > \gamma)\end{equation}

where $\mathbb{I}(\cdot)$ is the indicator function, and $\gamma$ is a correctness threshold set to 0.5. The rationale for this choice and sensitivity analysis of $\gamma$ are provided in Appendix \ref{sec:gamma}.
Using this binary classification, we identify the \textit{decisive error step} $t^*$ as the earliest failure point in the trajectory $\tau$:
\begin{equation}t^* = \min\{t \mid B_t = 0, 0 \leq t \leq T\}\end{equation}

The \textit{\afamethod} model is trained on the \textit{Who \& When Pro} dataset, which consists of annotated reasoning trajectories $\mathcal{D} = \{(\tau^{(j)}, \{l_t^{(j)}\}_{t=0}^{T_j})\}_{j=1}^M$, where $M$ denotes the total number of trajectories in the dataset. Each trajectory $\tau^{(j)}$ is paired with annotated binary labels $\{l_t^{(j)}\}_{t=0}^{T_j}$, indicating the correctness of every reasoning step from $t=0$ to $T_j$, the length of the $j$-th trajectory. More details about the dataset are provided in Section \ref{sec:whowhendataset2}.
The training objective is to minimize the Binary Cross-Entropy (BCE) loss:
\begin{equation}
\mathcal{L}_{\text{\textit{\afamethod}}} = -\frac{1}{M} \sum_{j=1}^{M} \sum_{t=0}^{T_j} \text{BCE}(c_t^{(j)}, l_t^{(j)})
\end{equation}

\subsection{Correction Submodule}
\label{sec:correction}

Once the failure points have been identified, our framework employs counterfactual reasoning \cite{bottou2013counterfactual,qin-etal-2019-counterfactual} to generate the \textit{corrected decisive error step}, which is essential for generating informative reflections that can guide future improvements.

Given the \textit{decisive error step} $t^*$ and \textit{decisive error agent} $i^*$, we generate a counterfactual action $\tilde{a}_{t^*}$ that represents a corrected version of the \textit{decisive error step} action $a_{t^*}$ :
\begin{equation}\tilde{a}_{t^*} = \mathcal{CR}(s_{t^*}, \tau_{<t^*}, p_{i^*})\end{equation}

where $\mathcal{CR}$ represents the counterfactual reasoning function that takes the current state, the trajectory history up to \textit{decisive error step} $t^*$, and the \textit{decisive error agent} profile $p_{i^*}$ to produce an alternative action to correct the decisive error.

To evaluate the quality of the counterfactual action $\tilde{a}_{t^*}$ we apply \textit{\afamethod} to compute a new correctness score:
\begin{equation}\tilde{c}_{t^*} = \afamethod_{\phi}(s_{t^*}, \tilde{a}_{t^*}, \tau_{<t^*})\end{equation}

A successful correction should result in $\tilde{c}_{t^*} > \gamma$, indicating that the counterfactual action $\tilde{a}_{t^*}$ successfully addresses the error and can be considered as the \textit{corrected decisive error step}. If $\tilde{c}_{t^*} \leq \gamma$, the correction is deemed unsuccessful, and the counterfactual reasoning process $\mathcal{CR}$ would be invoked again to generate a new alternative action until a satisfactory correction is achieved.

\subsection{Reflector Model}
\label{sec:reflection}

The reflector performs targeted reflection for the \textit{decisive error agent}, focusing on identifying the root causes of the \textit{decisive error step} and formulating a completely different, concise, high-level plan aimed at mitigating similar failures in future tasks. 
To enhance the quality of generated reflections, we further fine-tune the reflector using proximal policy optimization \citep{schulman2017proximal}.

\subsubsection{Instruction and Reflection Generation}

For a given failed trajectory $\tau$, we first use the diagnosis submodule to locate the \textit{decisive error step} action $a_{t^*}$, and identify the corresponding \textit{decisive error agent} profile $p_{i^*}$. After obtaining the \textit{corrected decisive error step} action $\tilde{a}_{t^*}$ from the correction submodule, we construct the input context $x = (\tau, p_{i^*}, a_{t^*}, \tilde{a}_{t^*})$ for the reflector. The reflector $\mathcal{M}_r$ then generates a targeted reflection $y_{i^*}$ for the \textit{decisive error agent} $i^*$:
\begin{equation}y^{i^*} = M_r(\tau, p_{i^*}, a_{t^*}, \tilde{a}_{t^*}) \end{equation}

\subsubsection{Rating Score of Reflections}
For a precise reflection $y_{k,e}^{i^*}$ generated by \textit{decisive error agent} $i^*$ in episode $e$ of task $k$, let $G_{k,e}$ denote the current episode return and $G_{k,e+1}$ represent the subsequent episode return that incorporates reflection $y_{k,e}^{i^*}$. Since the actor agents are based on frozen language models with low temperature settings \cite{shinn2023reflexion,yao2024retroformer,bo2024reflective}, the injected randomness that leads to return variation $\Delta G $ is primarily attributed to the reflection $y_{k,e}^{i^*}$. Hence, we define the rating score of reflection $y_{k,e}^{i^*}$ as:
\begin{equation}
r(y_{k,e}^{i^*}) = G_{k,e+1} - G_{k,e}
\end{equation}

\subsubsection{Optimization of the Reflector}

We optimize the reflector $\mathcal{M}_r$ through a three-stage \textit{RLHF} \cite{ouyang2022training} pipeline.
We first collect high-quality reflections with positive scores as demonstration examples and train a supervised reflector $\pi_{\text{SFT}}$ using Supervised Fine-Tuning (SFT) with standard maximum likelihood estimation:
\begin{equation}\mathcal{L}_{\text{SFT}}(\theta) = -\mathbb{E} \left[ \log \pi_{\theta}(y^{i^*} \mid x)  \right] \end{equation}

Taking construction expenses into account, instead of collecting pairwise responses for each input, we train a regression model to predict the rating scores of input prompts and reflection pairs. 
We then optimize the reward model $R_\phi$ by minimizing the Mean Square Error (MSE) loss:
\begin{equation}
\mathcal{L}_{\text{RM}}(\phi) = \mathbb{E} \left[ \left(R_\phi(x, y^{i^*}) - r\right)^2 \right] 
\end{equation}

Finally, we fine-tune the supervised reflector $\pi_{\text{SFT}}$ using \textit{PPO} with the trained reward model $R_\phi$. The reflector learns to generate reflections that maximize the expected reward by minimizing the following loss objective: 
\begin{equation}
\mathcal{L}_{\text{PPO}}(\theta) = - \mathbb{E} \left[ R_\phi(x, y) - \beta \log \frac{\pi_\theta^{\text{RL}}(y^{i^*} \mid x)}{\pi_{\text{SFT}}(y^{i^*} \mid x)} \right]
\end{equation}

where $\beta$ controls the strength of the KL divergence regularization term, ensuring that the fine-tuned model does not deviate too far from the reference model $\pi_{\text{SFT}}$.

%% file: 4.Experiment.tex
\section{Experiments}

\begin{figure*}[hbtp]
    \centering
    \includegraphics[width=0.95\linewidth]{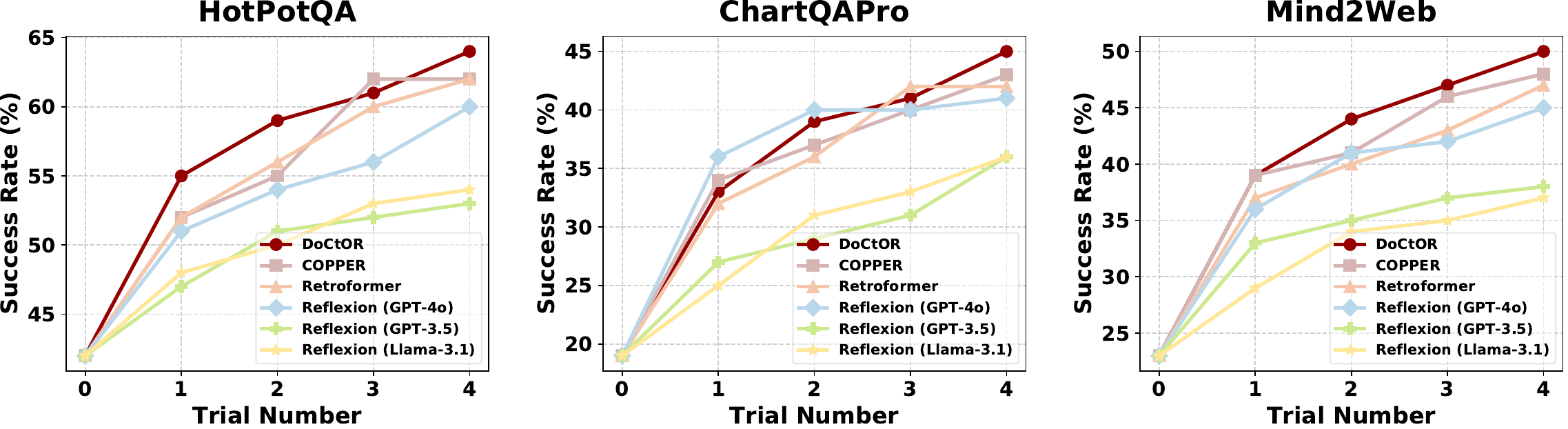}
    \caption{Performance comparison of \textit{\ourmethod} with baselines for \textit{self-reflection}.}
    \label{fig:mainresult}
\end{figure*}

\subsection{Datasets}

\paragraph{DoCtOR} We evaluate \textit{\ourmethod} on HotPotQA \cite{yang2018hotpotqa}, ChartQAPro \cite{masry2025chartqapro}, and Mind2Web \cite{deng2023mind2web} to evaluate the collaborative abilities of multi-agent systems in multi-hop question answering, chart data analysis, and website interaction: \textbf{(\Rmnum{1})} \textbf{HotPotQA} is a multi-hop question-answering dataset comprising 113k Wikipedia-based question-answer pairs, specifically designed to assess web retrieval capabilities by requiring multiple reasoning steps to derive answers. \textbf{(\Rmnum{2})} \textbf{ChartQAPro} is a comprehensive benchmark featuring 1,341 charts encompassing various chart types, including infographics and dashboards, alongside 1,948 questions spanning multiple-choice, hypothetical, and unanswerable formats. \textbf{(\Rmnum{3})} \textbf{Mind2Web} is a dataset designed for evaluating execution capabilities across diverse website domains through feasible interactions like clicking, selecting, and typing, based on high-level user instructions.

\label{sec:whowhendataset2}

\paragraph{ProFA} To develop and evaluate \textit{\afamethod}, we constructed a \textbf{\textit{Who \& When Pro}} dataset by extracting 1000, 500, and 1000 tasks from HotPotQA, ChartQAPro, and Mind2Web respectively, yielding 667, 379, and 711 failure logs. These logs underwent initial annotation by GPT-4o followed by thorough review by three researchers to identify \textit{decisive error agent} (who) and \textit{decisive error step} (when). We partitioned this dataset in a 7:1.5:1.5 ratio for training, validation, and testing. 
More details on the dataset construction and annotation process are outlined in Appendix \ref{sec:datacollectdetail}.

The \textbf{\textit{held-in}} dataset corresponds to the test set of our constructed \textit{Who \& When Pro} dataset and contains failure logs from the same domains used during the training of \textit{\afamethod}.
To assess \textit{\afamethod}'s generalization capabilities, we employed the \textit{Who \& When} dataset from \cite{zhang2025which} as our \textbf{\textit{held-out}} dataset. This held-out test set comprises 184 failure logs from multi-agent systems with fine-grained annotations about failure-responsible agents and decisive error steps, including two subsets: \textit{Algorithm-Generated} and \textit{Hand-Crafted}.

\subsection{Baselines}

\paragraph{DoCtOR}

We compare \textit{\ourmethod} with the following self-reflection baselines:
\textbf{(\Rmnum{1})} \textbf{Reflexion} \cite{shinn2023reflexion}: A prompt-based reflection method that generates verbal feedback from environmental signals to enhance task performance.
\textbf{(\Rmnum{2})}  \textbf{Retroformer} \cite{yao2024retroformer}: Originally designed for single-agent reflection using policy gradient optimization, we extend this approach to multi-agent systems by training a shared reflector with \textit{overall rewards} as supervision signals.
\textbf{(\Rmnum{3})} \textbf{COPPER} \cite{bo2024reflective}: A multi-agent adaptation of Retroformer that trains a shared reflector using \textit{counterfactual rewards} to assess the contribution of a single agent’s reflection within the system, alleviating the credit assignment problem.
Detailed model configurations and prompts for all baselines are provided in Appendix \ref{sec:baselinemodel} and Appendix \ref{sec:Reflexionprompts} .

\paragraph{ProFA}

We compare \textit{\afamethod} with four automated failure attribution baselines from \cite{zhang2025which} based on GPT-4o model:
\textbf{(\Rmnum{1})} \textbf{All-at-once}: An LLM analyzes the complete failure log to simultaneously identify the \textit{decisive error agent} and \textit{decisive error step}.
\textbf{(\Rmnum{2})} \textbf{Step-by-step}: An LLM examines the failure log step-by-step, terminating upon identifying the first erroneous step as the \textit{decisive error step} and its agent as the \textit{decisive error agent}.
\textbf{(\Rmnum{3})} \textbf{Binary search}: An LLM iteratively narrows down the failure log by half until isolating a single step, identifying as the \textit{decisive error step}.
\textbf{(\Rmnum{4})} \textbf{Random}: A naive baseline that randomly selects \textit{decisive error agent} and \textit{decisive error step}.

\subsection{Implementation Details}

\subsubsection{Collaboration Setting} 
We employ the Magnetic-One \cite{fourney2024magentic} multi-agent system to execute tasks from HotPotQA, ChartQAPro, and Mind2Web. Magnetic-One is a general multi-agent system designed to solve tasks such as software engineering, data analysis, scientific research, and web navigation. More details regarding Magnetic-One's architecture, multi-agent team configurations across different datasets, and response formats are outlined in Appendix \ref{sec:Collaboration}.

\subsubsection{Training} 
To develop \afamethod, we conduct full-parameter fine-tuning on Qwen3-1.7B. Furthermore, we fine-tune Llama-3.1-8B-Instruct with \textit{LoRA} \cite{hu2022lora} as the reflector and use GPT-2 as the regression reward model, following the setup in  \cite{bo2024reflective}. More training details are provided in Appendix \ref{sec:trainingdetail}.

\subsubsection{Evaluation}

To evaluate \textit{self-reflection} methods including \textit{\textbf{\ourmethod}}, we adopt the \textbf{Success Rate} metric following prior work \cite{shinn2023reflexion,yao2024retroformer,bo2024reflective}. A task is considered successful if the generated answer sufficiently matches the ground-truth answer. To quantify this match, we use the F1 score, which enables soft matching rather than exact binary comparison. Details of the F1 score computation are provided in Appendix \ref{sec:RewardFunction}. Due to computational constraints and consistent with prior work \cite{shinn2023reflexion,yao2024retroformer,bo2024reflective}, we construct the test set by randomly sampling 100 tasks from each dataset.

For assessing \textit{automated failure attribution} methods including \textit{\textbf{\afamethod}}, we employ two metrics from \cite{zhang2025which}: \textbf{(\Rmnum{1})} \textbf{Agent-Level Accuracy}, the percentage of correctly identified \textit{decisive error agent}; and \textbf{(\Rmnum{2})} \textbf{Step-Level Accuracy}, the percentage of correctly identified \textit{decisive error step}.

\begin{table*}[]
\centering
\resizebox{0.95\textwidth}{!}{
\begin{tabular}{ccccccc}
\toprule
\multirow{2}{*}[-2ex]{\textbf{Baseline}} & \multirow{2}{*}[-2ex]{\textbf{Metric}} & \multicolumn{3}{c}{\textbf{Held-in Dataset}} & \multicolumn{2}{c}{\textbf{Held-out Dataset}} \\ \cmidrule(l){3-5}  \cmidrule(l){6-7} 
 &  & \multicolumn{1}{c}{\textbf{HotPotQA}} & \multicolumn{1}{c}{\textbf{ChartQAPro}} & \multicolumn{1}{c}{\textbf{Mind2Web}} & \multicolumn{1}{c}{\textbf{\begin{tabular}[c]{@{}c@{}}Algorithm-\\ Generated\end{tabular}}} & \multicolumn{1}{c}{\textbf{\begin{tabular}[c]{@{}c@{}}Hand-\\ Crafted\end{tabular}}} \\ \midrule
\multirow{2}{*}{\textbf{Random}} & \textbf{Agent-Level Accuracy} & 0.37 & 0.29 & 0.54 & 0.28 & 0.20 \\
 & \textbf{Step-Level Accuracy} & 0.06 & 0.09 & 0.02 & 0.18 & 0.04 \\ \midrule
\multirow{2}{*}{\textbf{All-at-Once}} & \textbf{Agent-Level Accuracy} & 0.54 & 0.55 & 0.19 & 0.49 & 0.51 \\
 & \textbf{Step-Level Accuracy} & 0.28 & 0.45 & 0.23 & 0.10 & 0.05 \\ \midrule
\multirow{2}{*}{\textbf{Step-by-Step}} & \textbf{Agent-Level Accuracy} & 0.45 & 0.42 & 0.51 & 0.28 & 0.42 \\
 & \textbf{Step-Level Accuracy} & 0.20 & 0.42 & 0.12 & 0.17 & 0.11 \\ \midrule
\multirow{2}{*}{\textbf{Binary Search}} & \textbf{Agent-Level Accuracy} & 0.37 & 0.38 & 0.74 & 0.32 & 0.38 \\
 & \textbf{Step-Level Accuracy} & 0.18 & 0.24 & 0.02 & 0.12 & 0.09 \\ \midrule
\multirow{2}{*}{\textbf{Our}} & \textbf{Agent-Level Accuracy} & 0.85 & 0.82 & 0.79 & 0.53 & 0.55 \\
 & \textbf{Step-Level Accuracy} & 0.53 & 0.78 & 0.42 & 0.38 & 0.20 \\ \bottomrule
\end{tabular}}
    \caption{Performance comparison of \textit{\afamethod} with baselines for \textit{automated fault attribution}.}
    \label{tab:promainresult}
\end{table*}

\begin{figure*}[t]
    \centering
    \includegraphics[width=0.95\linewidth]{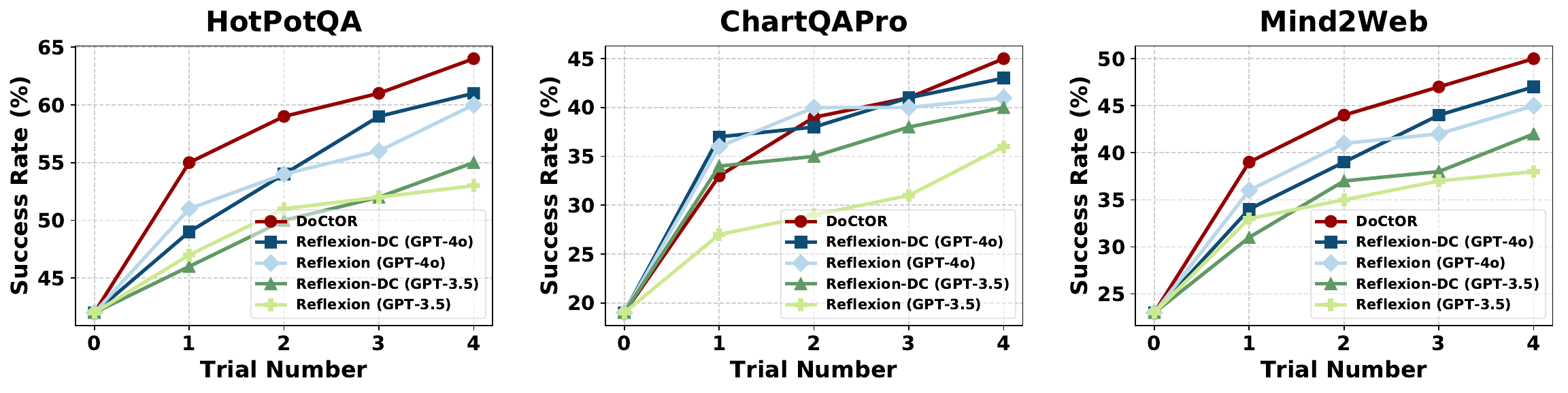}
    \caption{Generalizability of the \textit{diagnose-then-correct} reflection paradigm. We evaluate the paradigm on the prompt-based Reflexion method to verify its generalizability beyond our fine-tuning-based implementation.}
    \label{fig:Generalizability}
\end{figure*}

\subsection{Main Results}

We evaluated the \textit{\ourmethod} framework across HotPotQA, ChartQAPro, and Mind2Web datasets over five trials, as illustrated in Figure \ref{fig:mainresult}. Our analysis reveals that \textit{\ourmethod} exhibited superior reflection capabilities compared to all baseline methods including Reflexion, Retroformer, and COPPER. By leveraging \textit{\afamethod} to localize the \textit{decisive error agent} and \textit{decisive error step}, then employing counterfactual reasoning to generate the \textit{corrected decisive error step}, \textit{\ourmethod} provides targeted auxiliary information that enables the \textit{decisive error agent} to accurately identify failure causes and formulate more effective improvement strategies. Consequently, \textit{\ourmethod} achieved substantial improvements of 22\%, 26\%, and 27\% over initial success rates on HotPotQA, ChartQAPro, and Mind2Web datasets, respectively, demonstrating the effectiveness of our \textit{diagnose-then-correct} paradigm. Additional experiments on robustness and scalability across different model families and sizes are reported in Appendix \ref{sec:rubost_model}.

\subsection{Performance of ProFA}

As shown in Table \ref{tab:promainresult}, \textit{\afamethod} demonstrated robust performance, achieving approximately 80\% \textit{agent-level accuracy} on held-in datasets and 50\% on held-out datasets, while \textit{step-level accuracy} reached 53\%, 78\%, and 42\% on held-in datasets and 38\% and 20\% on held-out datasets respectively. 

For \textit{held-in datasets}, the performance improvements over baselines were particularly notable on the HotPotQA dataset, where \textit{\afamethod} achieved 31\% improvement in agent-level accuracy and 25\% in step-level accuracy compared to all-at-once, and on the ChartQAPro dataset with 27\% improvement in agent-level accuracy and 33\% in step-level accuracy compared to all-at-once. 
For \textit{held-out datasets}, \textit{\afamethod} maintained its advantage with 4\% improvements in agent-level accuracy over all-at-once, while step-level accuracy improved by 20\% over random on Algorithm-Generated dataset and 9\% over step-by-step on Hand-Crafted dataset. 

\begin{figure}[!h]
    \centering
    \includegraphics[width=\linewidth]{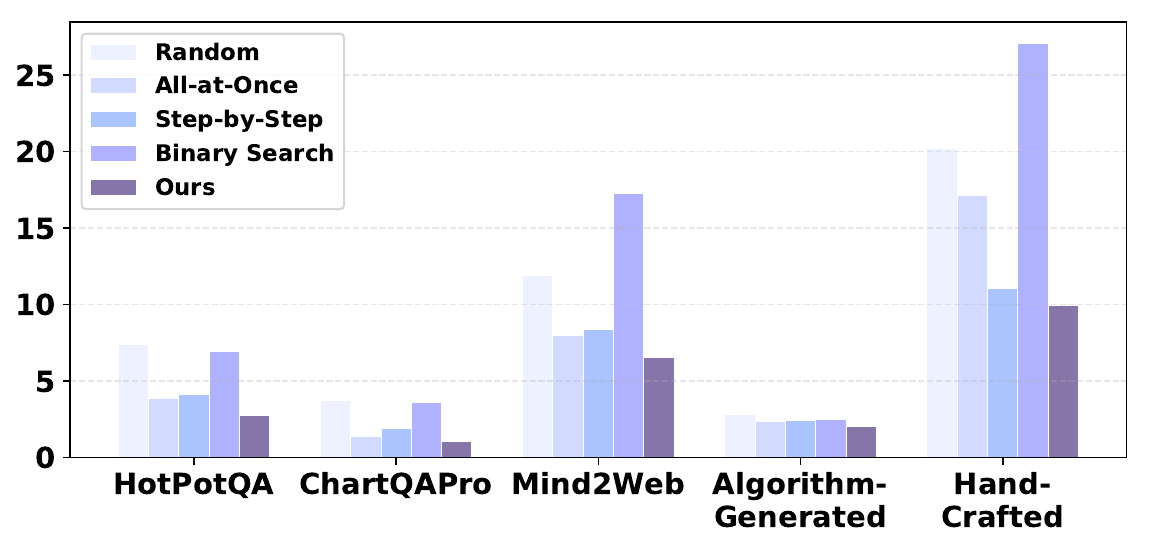}
    \caption{The absolute distance between the predicted and actual \textit{decisive error step}.}
    \label{fig:error}
\end{figure}

Additionally, as illustrated in Figure \ref{fig:error}, \textit{\afamethod} consistently achieved the smallest absolute distance between predicted and actual \textit{decisive error step} across all datasets.

\begin{figure*}[t]
    \centering
    \includegraphics[width=0.95\linewidth]{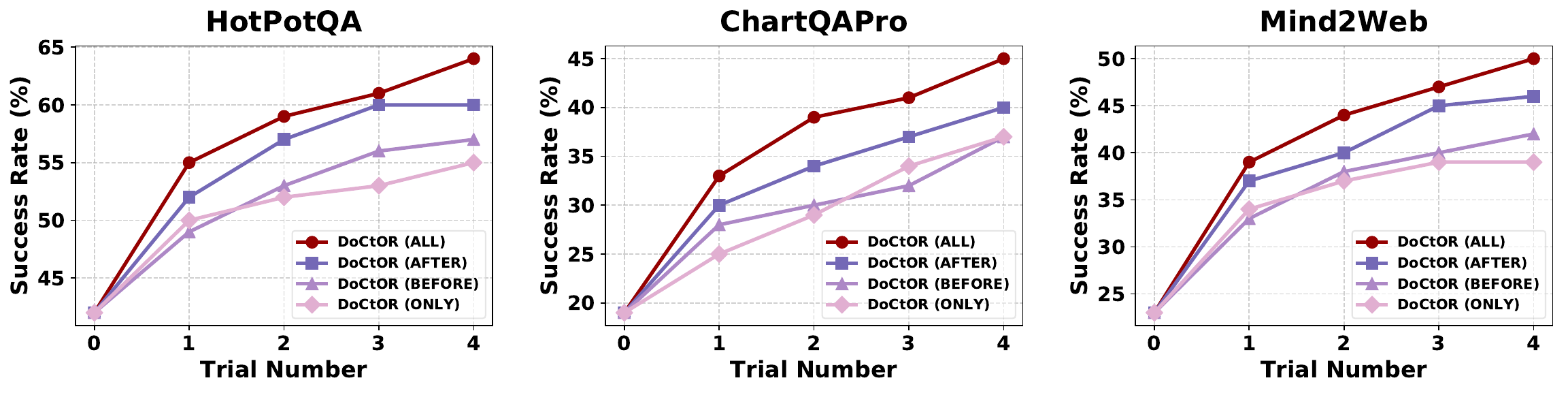}
    \caption{Effect of failure trajectory scope on reflection quality. We investigate the impact of different scopes of failed reasoning trajectories on reflection quality by comparing four strategies: the \textit{full} failure trajectory, steps \textit{after} the decisive error step, steps \textit{before} the decisive error step, and \textit{only} the decisive error step. }
    \label{fig:Context}
\end{figure*}

\begin{figure*}[t]
    \centering
    \includegraphics[width=0.95\linewidth]{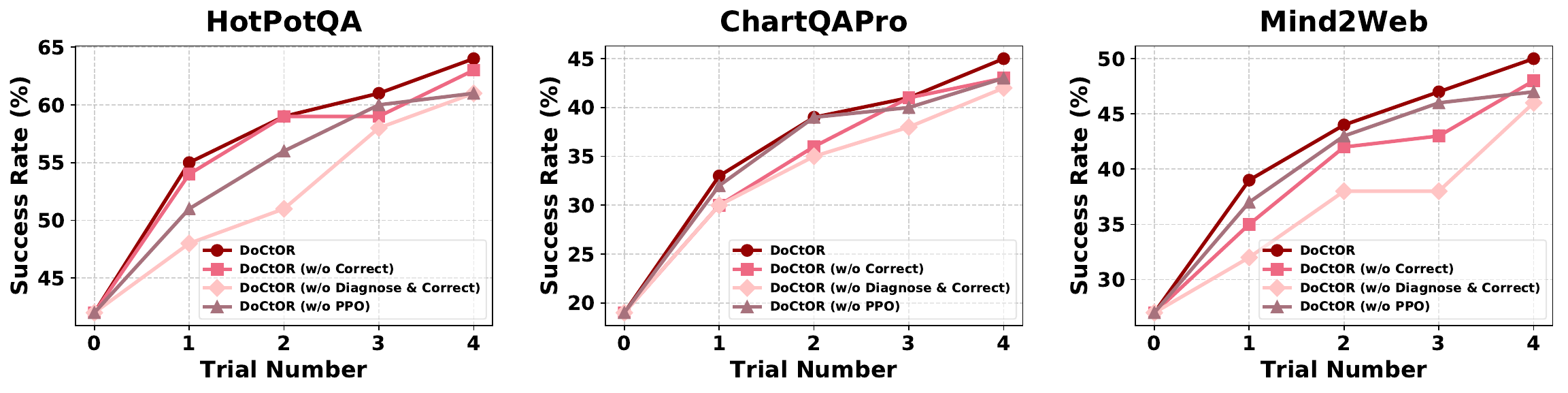}
    \caption{Ablation study. Performance comparison of \textit{\ourmethod} with variants that remove the correction submodule (w/o correct), both diagnosis and correction submodules (w/o diagnose \& correct), or proximal policy optimization for the reflector (w/o PPO).}
    \label{fig:Ablation}
\end{figure*}

\subsection{Generalizability of Diagnose-then-Correct Reflection Paradigm}

We evaluate the generalizability of our \textit{diagnose-then-correct} paradigm by applying it to the prompt-based Reflexion method with GPT-3.5 and GPT-4o as base models, as shown in Figure \ref{fig:Generalizability}. The \textit{diagnose-then-correct} paradigm yields consistent but modest gains for GPT-4o-based Reflexion across all datasets, while achieving substantial improvements for GPT-3.5-based Reflexion on the ChartQAPro and Mind2Web datasets. 
The pronounced improvement with GPT-3.5 suggests that models with weaker intrinsic reasoning capabilities benefit more from the structured diagnostic and corrective guidance provided by our paradigm.

\subsection{Effect of Failure Trajectory Scope on Reflection Quality}

As shown in Figure \ref{fig:Context}, providing steps \textit{after} the decisive error step yields comparable performance to providing a complete multi-agent trajectory.
This indicates that steps \textit{after} the decisive error step contain sufficient information for generating high-quality reflection, as the reflector can identify fundamental issues by analyzing error cascades. These insights have significant practical implications for low-resource environments, demonstrating that effective reflection can be achieved by strategically selecting reasoning steps after the decisive error step, thereby optimizing computational efficiency while maintaining reflection quality.

\subsection{Ablation Study}

We conduct an ablation study to assess the contribution of each component in the \ourmethod framework. As illustrated in Figure \ref{fig:Ablation}, we remove the correction submodule, both diagnosis and correction submodules, and PPO for the reflector. The results demonstrate that all components are crucial for optimal performance, as removing any of them leads to consistent performance drops across all datasets. In particular, removing both diagnosis and correction submodules causes the largest degradation, underscoring the importance of the \textit{diagnose-then-correct} paradigm.

%% file: 6.Conclusion.tex
\section{Conclusion}

In this work, we introduced \ourmethod, a novel reflection framework that significantly enhances multi-agent collaboration. Our key contribution lies in recognizing that failures typically stem from a decisive error agent leading the task astray, while others merely fulfill their regular duties. Through automated failure attribution and targeted reflection, \ourmethod prevents memory contamination for regular-behaving agents while achieving superior reflection quality, paving the way for more effective multi-agent collaboration systems.

\section*{Limitations}
While the \ourmethod framework achieves promising results, several areas remain for future exploration. First, our current implementation focuses on task-oriented multi-agent collaborations with clear success criteria. Extending the framework to open-ended creative tasks or scenarios with subjective evaluation metrics presents an interesting direction for future research. Second, we primarily evaluated \ourmethod on English-language datasets and tasks. The framework's performance across multilingual settings and culturally diverse collaborative scenarios would benefit from further investigation to establish broader applicability. Finally, our experiments primarily involve relatively small multi-agent teams. Investigating the framework's scalability with larger multi-agent teams would provide valuable insights for deploying \ourmethod in more complex collaborative scenarios.

\section*{Acknowledgments}
This work was supported by the National Natural Science Foundation of China under Grant Nos. 62441230, 62172425, and 62672497, the Scientific Research Innovation Capability Support Project for Young Faculty under Grant No. SRICSPYF-ZY2025001, and the Fundamental Research Funds for the Central Universities and the Research Funds of Renmin University of China under Grant No. 22XNKJ04.

%% file: 7.Appendix.tex
\clearpage

\section{Related Work}

\subsection{LLM-based Multi-Agent Systems}

Recent years have witnessed substantial attention toward Large Language Model (LLM)-based agent systems within the artificial intelligence domain \cite{wang2024survey,zhao2024expel}. Building upon single-agent architectures, LLM-based multi-agent systems have experienced rapid advancement, demonstrating significant progress in complex task decomposition and world simulation capabilities. The growing interest in these systems stems from their capacity to handle intricate multi-step tasks while dynamically interacting with diverse environments, making them particularly suitable for real-world applications \cite{li2023camel}. Multi-agent systems have been increasingly explored across various domains, including software engineering \cite{qian2024chatdev,hong2024metagpt,wang2025openhands}, drug discovery \cite{gottweis2025towards,swanson2024virtual}, scientific simulation \cite{tang2023medagents,ghafarollahi2025sciagents}, and general-purpose intelligence \cite{fourney2024magentic,wu2024autogen,chen2023agentverse,li2023camel}.

However, recent studies \cite{wang2024rethinking,zhang2025if,pan2025why} reveal that multi-agent deliberation methods do not consistently outperform simpler single-agent baselines, even with additional computational resources for reasoning. These findings suggest the need for optimized collaborative mechanisms to enhance multi-agent problem-solving capabilities significantly.

\subsection{Self-Reflection of Large Language Models}

Reflection mechanisms have gained significant attention as they employ self-reflection feedback to provide specific improvement directions for agents. These approaches enable agents to learn from previous errors, avoid recurring mistakes, and perform better in subsequent attempts. Early investigations focused on optimizing responses based on singular feedback \cite{madaan2023self,chen2024teaching} or comparative analysis between multiple models \cite{du2024improving,zhang2024self}, but failed to develop comprehensive task understanding from past experiences. 
Later work \cite{shinn2023reflexion} examined previous trajectories and environmental rewards to generate reflections, incorporating these insights into subsequent contexts. While this approach enables iterative enhancement, its effectiveness depends largely on the model's inherent reflective capabilities. To address this limitation, \cite{yao2024retroformer} proposed Retroformer, which approximates reflection rewards through differentials between consecutive outcomes and trains a reflector via policy optimization. More recently, \cite{bo2024reflective} extended this methodology to multi-agent systems with COPPER, employing counterfactual rewards to evaluate reflection quality for each agent and determine whether reflections merit inclusion in agent memory.

However, existing reflection frameworks overlook a critical nuance: when tasks fail, not every agent bears responsibility for the failure. More commonly, a specific agent leads the task astray while others simply fulfill their normal duties. To address this limitation, we propose \ourmethod, a novel reflection framework that enhances multi-agent collaboration through targeted diagnostic and corrective mechanisms.

\section{Implementations Details}

\subsection{Baseline Models}
\label{sec:baselinemodel}

To ensure fair comparison, all baselines employ GPT-4o-mini for their action modules, consistent with our \ourmethod framework's action module. For the reflection modules, we adopt different model configurations based on each method's training requirements. Fine-tuning-based approaches, including Retroformer and COPPER, utilize Llama-3.1-8B-Instruct as the base model, matching the foundation model of our \ourmethod reflector. For the prompt-based Reflexion method, we evaluate performance across three base models: Llama-3.1-8B-Instruct, GPT-4o, and GPT-3.5, providing comprehensive comparison results.

\subsection{Training}
\label{sec:trainingdetail}

\subsubsection{DoCtOR}
\paragraph{Data Collection}
To fine-tune our reflector, we extracted 1000, 500, and 1000 tasks from each dataset respectively, with a maximum of 3 trials per task, yielding 3565 reflection instances. 

\paragraph{Training Details} We employed Low-Rank Adaptation (LoRA) for parameter-efficient fine-tuning and implemented Reinforcement Learning from Human Feedback (RLHF) through HuggingFace's trl package. Our training pipeline incorporated a three-stage approach: supervised fine-tuning, reward modeling, and policy optimization. For supervised fine-tuning, we conducted a grid search including epochs {1, 2, 3, 4}, batch sizes {8, 16, 32, 64, 128}, and learning rates {1e-4, 2e-4, 3e-4, 4e-4, 5e-4} using a validation set of 100 instances. The reward model was trained with a learning rate of 1e-5, 3 epochs, and a batch size of 16. For PPO training, we utilized lower learning rates {1e-5, 2e-5, 3e-5, 4e-5, 5e-5} to ensure stable convergence. All models were fine-tuned using 4-bit quantized LoRA adapters, resulting in efficient parameter utilization (0.015\%-0.06\% of base model parameters).

\subsubsection{ProFA}

\paragraph{Data Collection}

\label{sec:whowhendataset}

\label{sec:datacollectdetail}
To develop and evaluate \textit{\afamethod}, we constructed a \textbf{\textit{Who \& When Pro}} dataset through a systematic data collection and annotation process. We extracted 1000, 500, and 1000 tasks from HotPotQA, ChartQAPro, and Mind2Web, respectively, yielding 667, 379, and 711 failure logs from unsuccessful multi-agent collaboration attempts.
The annotation process involved a two-stage approach to ensure data quality. 

First, we conducted initial automated annotation using GPT-4o to provide preliminary identification of \textit{decisive error agents} (who) and \textit{decisive error steps} (when) within each failure log. 
Subsequently, three experienced researchers independently reviewed and refined these annotations. Each researcher examined the multi-agent interaction traces, analyzing causal relationships between agent actions and task failures to determine the precise moment when the decisive error occurred and which agent was responsible. 
The three researchers achieved a Fleiss' kappa score of 0.83 during the annotation process, indicating substantial agreement, which falls into the "almost perfect agreement" range (0.81-1.00) on the standard Fleiss' kappa interpretation scale.

\paragraph{Training Details} We fine-tuned a Qwen3-1.7B model using a token classification approach. We employed the PRMTrainer from \textit{trl} with the following hyperparameters: learning rate of 5e-5, weight decay of 0.01, batch size of 8, maximum sequence length of 2048, and 3 training epochs. We implemented a warmup ratio of 0.1 with gradient accumulation steps set to 1.

\subsection{Evaluation}
\label{sec:RewardFunction}

Our framework employs the F1 score to evaluate generated answer quality by balancing precision and recall. The calculation process involves normalizing both prediction and ground truth responses. For categorical responses (yes, no, or no answer), the function assigns zero metrics when predictions do not match the ground truth. For standard text responses, the algorithm tokenizes the normalized texts and identifies overlapping tokens. Precision is calculated as the ratio of common tokens to prediction tokens, while recall represents the ratio of common tokens to ground truth tokens. The F1 score is computed as:

\begin{equation}F1 = \frac{2 \times precision \times recall}{precision + recall}\end{equation}

The Mind2Web benchmark, while ultimately evaluated on task-level execution correctness, presents challenges analogous to structured question-answering tasks, requiring agents to generate web navigation instructions in a specific format (e.g., "Answer: A/B/C/D Action:CLICK/SELECT/TYPE Value:XXX"). We selected the F1 score as our reward function because it effectively balances precision and recall in structured outputs, assessing both the accuracy of included elements and the comprehensiveness of responses. This approach provides more fine-grained feedback than binary task-level execution correctness metrics alone. Our preliminary experiments confirm this benefit, showing approximately 10.2\% performance improvement when using F1-based rewards compared to binary task-level execution correctness signals.

\subsection{Reproducibility}

All experiments were conducted on two NVIDIA A100-80G GPUs. To ensure experimental rigor, we configured different temperature settings for the base models in each module. For the action module's base model, we set the temperature to 0 and top-p to 1 following prior work \cite{shinn2023reflexion,yao2024retroformer,bo2024reflective}, thereby isolating the randomness inherent in language model generation from the effects of reflections. In contrast, for the reflection module's base model, we employed a temperature of 0.8 to encourage more creative and diverse reflection generation.

\section{Collaboration Settings}
\label{sec:Collaboration}

\subsection{Magentic-One Multi-Agent Framework}
Our implementation utilizes the Magentic-One framework, which employs a hierarchical multi-agent structure for complex task solving. The system comprises several specialized agents working in coordinated collaboration:

\begin{itemize}
    \item \textbf{Orchestrator}: Functions as the central coordinator responsible for task decomposition, strategic planning, and adaptive management of the agent workflow. The Orchestrator monitors execution progress and delegates subtasks to specialized agents.
    \item \textbf{WebSurfer}: Specializes in web-based information retrieval and browser interaction. This agent navigates URLs, performs searches, interacts with webpage elements (clicking, typing), and extracts relevant content through the browser's accessibility tree and set-of-marks prompting techniques.
    \item \textbf{Coder}: Focuses on code generation, data analysis, and artifact creation. This agent processes information collected by other agents, implements algorithmic solutions, and generates computational outputs required for task completion.
    \item \textbf{ComputerTerminal}: Provides system-level access for executing commands, running programs, and installing necessary libraries. This agent enables the framework to interact with the operating environment and execute computational processes.
\end{itemize}

\subsection{Dataset-Specific Collaboration Settings}
We tailored our multi-agent configurations to address the unique requirements of three datasets:
\begin{itemize}
    \item \textbf{HotPotQA}: For multi-hop question answering tasks, we utilized Magentic's predefined agent ensemble consisting of the Orchestrator, WebSurfer, FileSurfer, Coder, and Terminal agents. This configuration effectively supports the information retrieval and reasoning chains necessary for complex question answering.
    \item \textbf{ChartQAPro}: To address chart-based analytical queries, we extended the base configuration by implementing a custom ImageAgent with specialized capabilities for visual content interpretation. We developed two complementary tools for image question answering and visual description generation. The resulting team composition (Orchestrator, ImageAgent, Coder, and Terminal) enables comprehensive chart data analysis.
    \item \textbf{Mind2Web}: For website interaction tasks, we developed a specialized WebAgent defined as a helpful assistant with expertise in website design, navigation, and task execution. This agent works in conjunction with the Orchestrator, Coder, and Terminal agents to navigate complex web interfaces and execute multi-step interactions.
\end{itemize}

\subsection{Response Format}

\begin{figure}[H]
    \centering
    \includegraphics[width=0.85\linewidth]{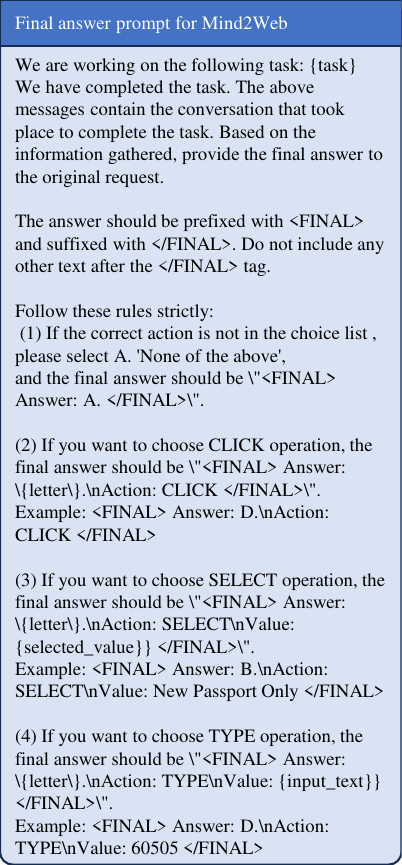}
    \caption{The final answer prompt for Mind2Web.}
    \label{fig:final_mind2web}
\end{figure}

\begin{figure}[H]
    \centering
    \includegraphics[width=0.85\linewidth]{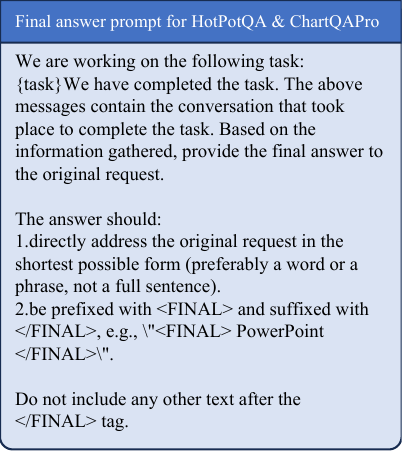}
    \caption{The final answer prompt for HotPotQA \& ChartQAPro
.}
    \label{fig:final_hotpot}
\end{figure}

To facilitate systematic evaluation and enhance analytical consistency across datasets, we implemented standardized response formats. For HotPotQA and ChartQAPro datasets, we developed a structured answer template, as illustrated in Figure \ref{fig:final_hotpot}. For the Mind2Web dataset, we established a more structured format that normalizes responses into a combination of Answer, Action, and Value components, also depicted in Figure \ref{fig:final_mind2web}.

\section{Additional Experiments}

\subsection{Effect of Test Set Size}

\label{sec:testset-size}

To comprehensively evaluate the scalability and consistency of our method, we conducted extensive experiments across different test set sizes on the HotPotQA benchmark. 

\begin{figure}[H]
    \centering
    \includegraphics[width=0.8\linewidth]{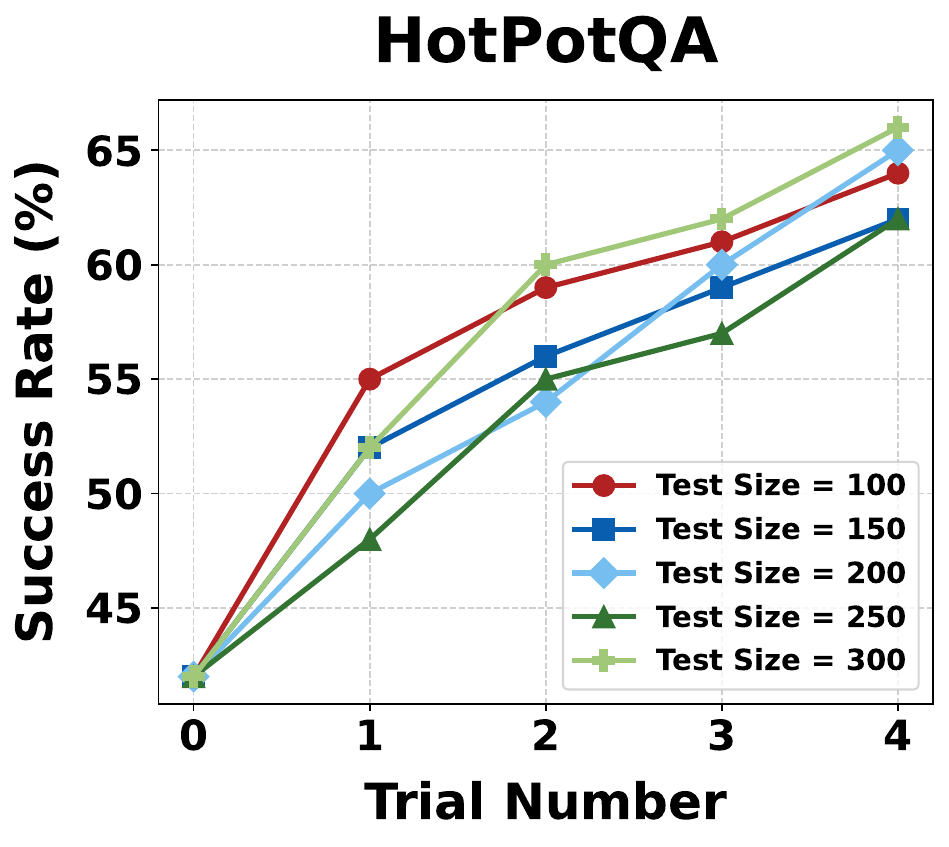}
    \caption{Performance on HotPotQA under different Test Set Sizes.}
    \label{fig:HotPotQA-trials}
\end{figure}

As shown in Figure \ref{fig:HotPotQA-trials}, our method consistently improves performance across all test set sizes throughout the trials. The stability of this improvement across varying scales indicates that the method's effectiveness is robust and generalizes well to larger evaluation sets.

\subsection{Robustness Across Model Families and Sizes}
\label{sec:rubost_model}
To evaluate scalability and robustness, we tested our framework with different model families and sizes.

\begin{figure}[H]
    \centering
    \includegraphics[width=0.8\linewidth]{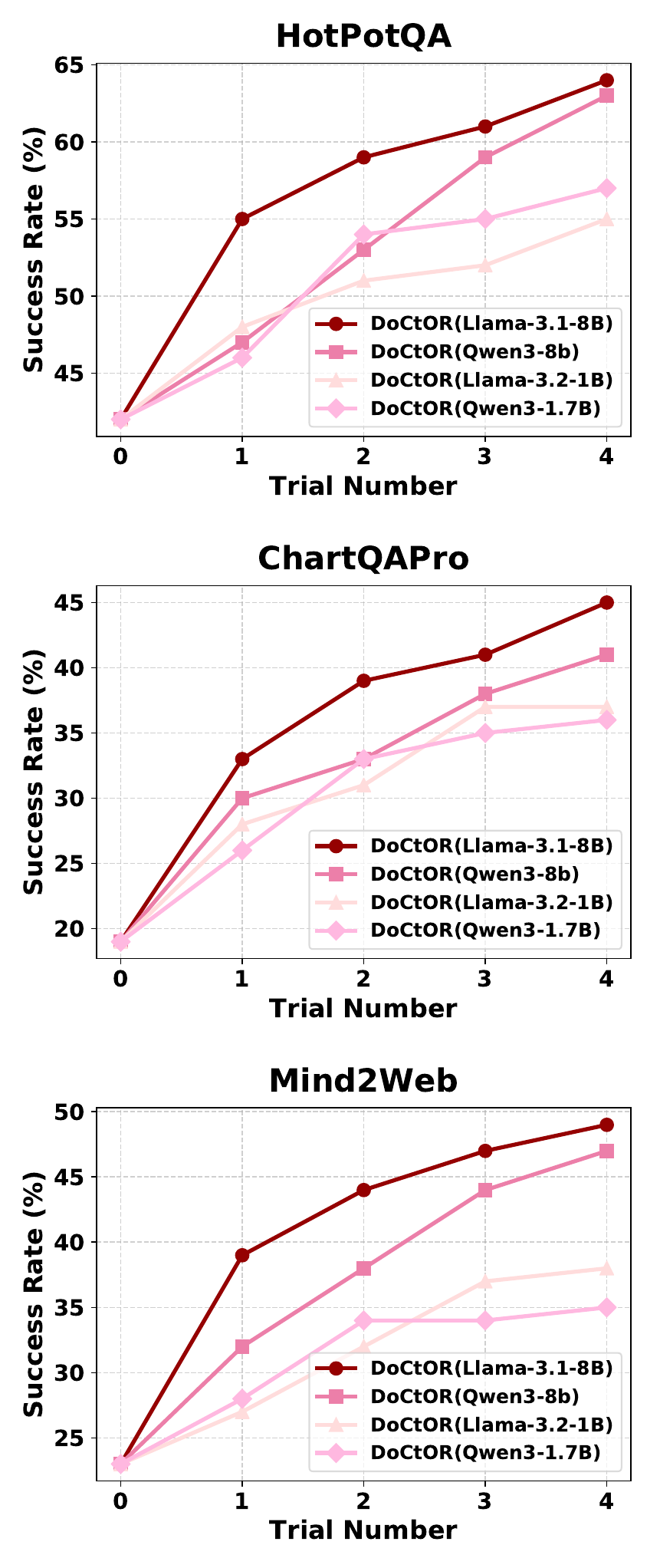}
    \caption{Robustness Across Model Families and Sizes.}
    \label{fig:Robustness_model}
\end{figure}

As illustrated in Figure \ref{fig:Robustness_model}, the results first validate our choice of Llama-3.1-8B while also demonstrating that our framework consistently improves performance across diverse model families and sizes, with larger models (8B) generally achieving better results than smaller ones (1-2B). 

\subsection{Effect of Correctness Threshold $\gamma$}

\label{sec:gamma}

The parameter $\gamma$ is set to 0.5 based on standard practices in PRM implementations within widely-used frameworks like TRL and OpenRLHF, which use 0.5 as the threshold for evaluating step correctness. 

Furthermore, we conducted extensive experiments to examine how the correctness threshold $\gamma$ affects final reflection performance.

\begin{figure}[H]
    \centering
    \includegraphics[width=0.8\linewidth]{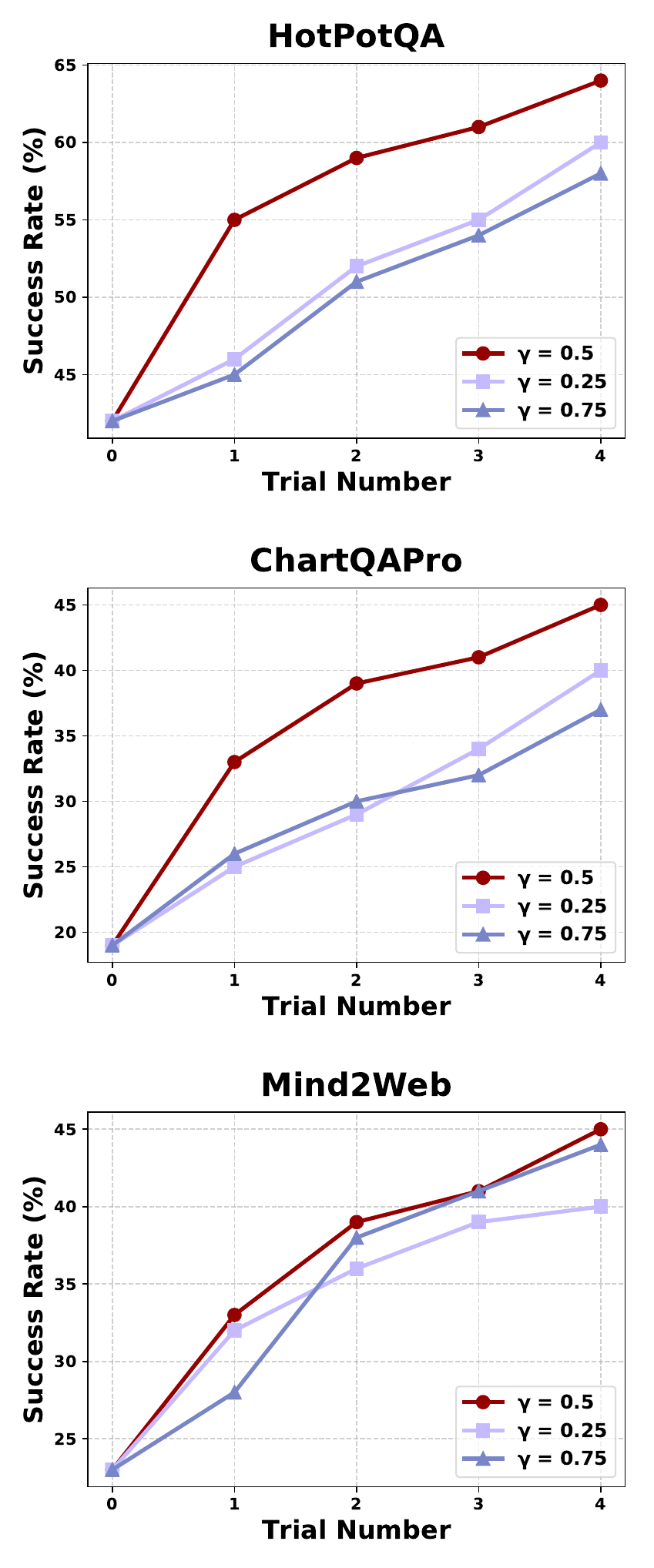}
    \caption{Effect of Correctness Threshold $\gamma$.}
    \label{fig:Hyperparameter}
\end{figure}

As shown in Figure \ref{fig:Hyperparameter}, the results demonstrate that while $\gamma$ = 0.5 generally yields the best performance across datasets, our method shows robust improvement patterns regardless of the specific threshold chosen.

\section{Prompts when Deploying DoCtOR}
\label{sec:DoCTORprompts}

\begin{figure}[H]
    \centering
    \includegraphics[width=\linewidth]{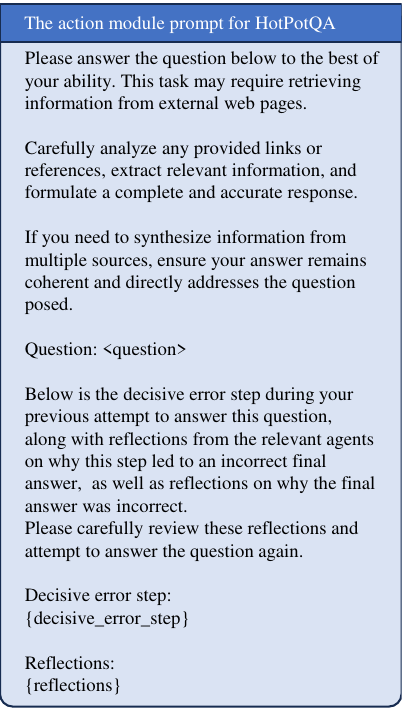}
    \caption{The action module prompt for HotPotQA when deploying \ourmethod.}
    \label{fig:gener}
\end{figure}

\vspace{-6cm}

\begin{figure}[t]
    \centering
    \includegraphics[width=\linewidth]{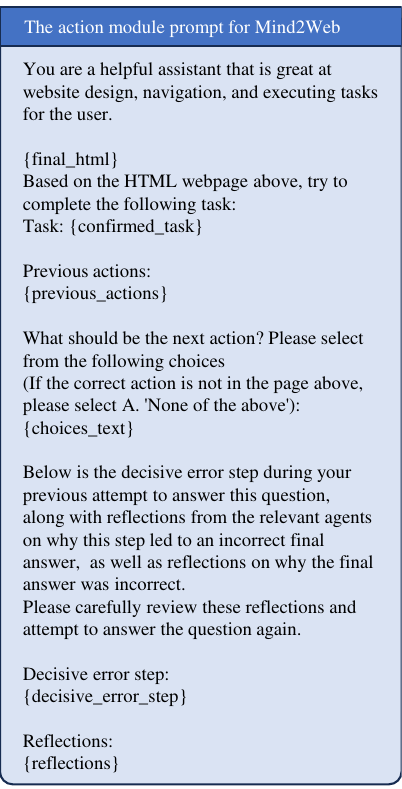}
    \caption{The action module prompt for Mind2Web when deploying \ourmethod.}
    \label{fig:gener}
\end{figure}

\begin{figure}[t]
    \centering
    \includegraphics[width=\linewidth]{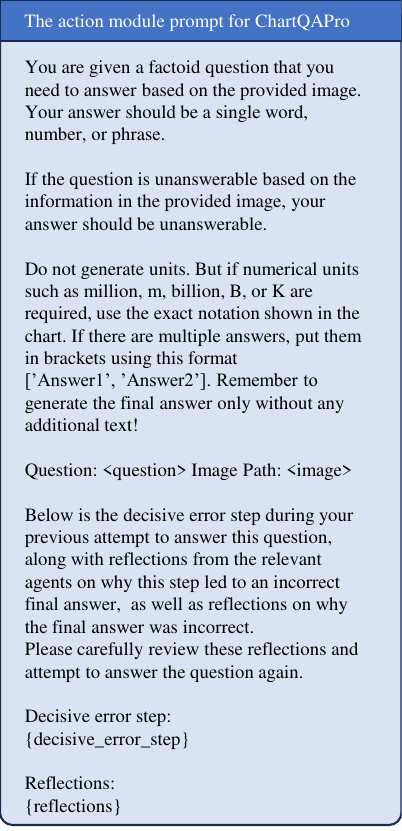}
    \caption{The action module prompt for ChartQAPro when deploying \ourmethod.}
    \label{fig:gener}
\end{figure}

\begin{figure}[]
    \centering
    \includegraphics[width=0.9\linewidth]{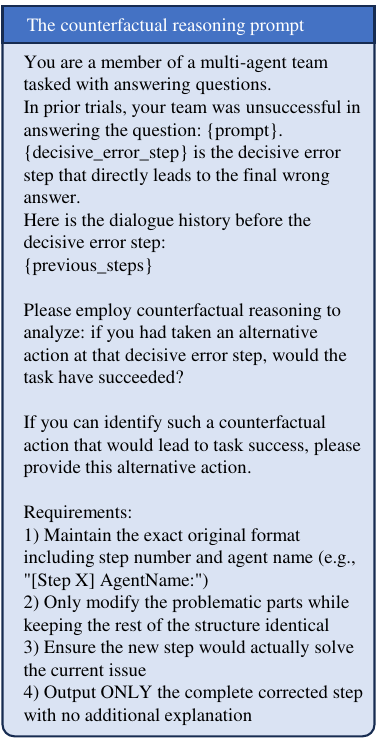}
    \caption{The counterfactual reasoning prompt when deploying \ourmethod.}
    \label{fig:gener}
\end{figure}

\begin{figure*}[]
    \centering
    \includegraphics[width=0.9\linewidth]{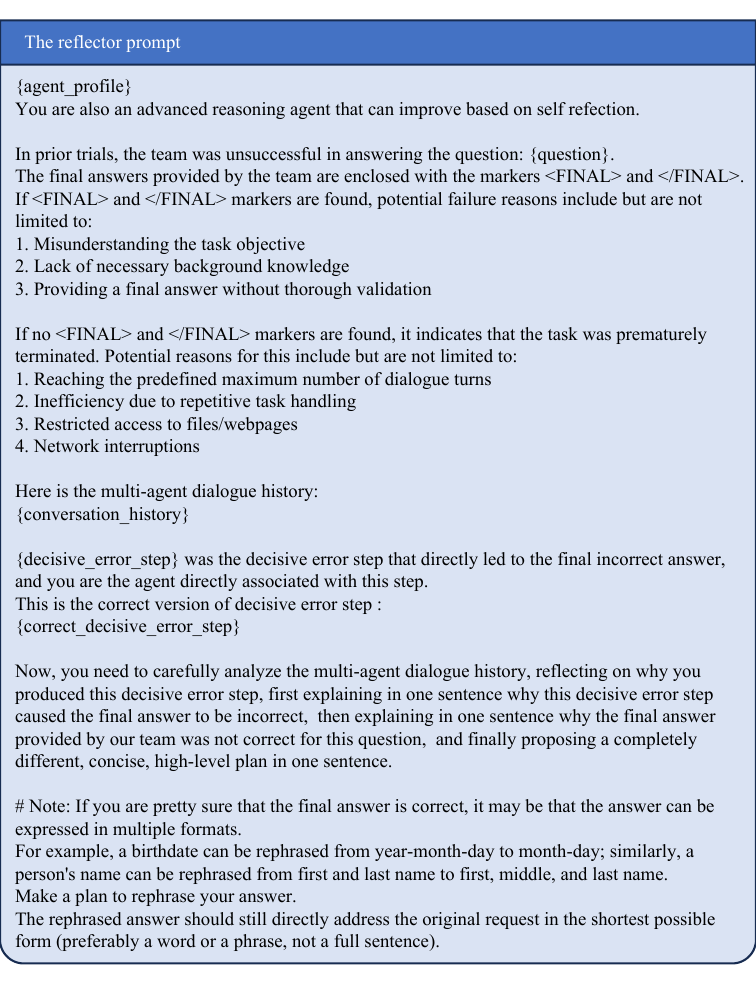}
    \caption{The reflector prompt when deploying \ourmethod.}
    \label{fig:gener}
\end{figure*}

\clearpage
\section{Prompts when Deploying Reflexion, Retroformer, and COPPER}
\label{sec:Reflexionprompts}

\begin{figure}[H]
    \centering
    \includegraphics[width=0.9\linewidth]{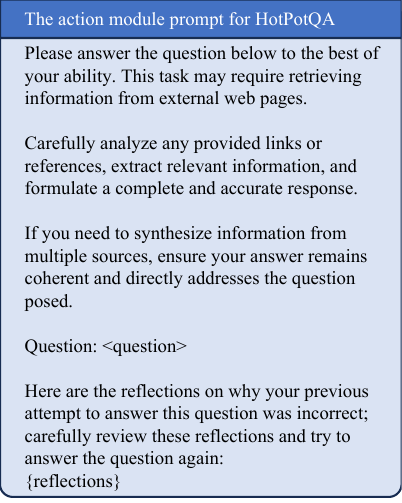}
    \caption{The action module prompt for HotPotQA when deploying Reflexion, Retroformer, and COPPER.}
    \label{fig:gener}
\end{figure}

\begin{figure}[H]
    \centering
    \includegraphics[width=0.9\linewidth]{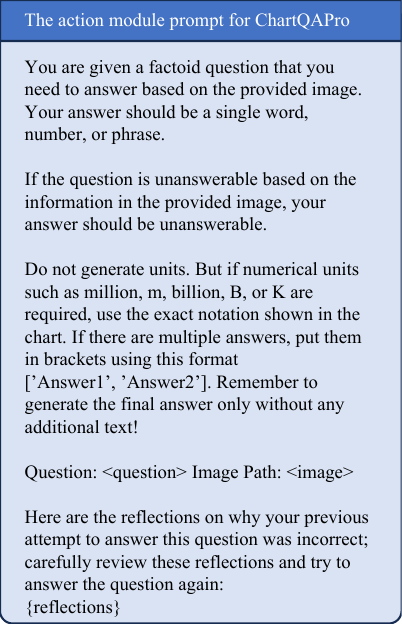}
    \caption{The action module prompt for ChartQAPro when deploying Reflexion, Retroformer, and COPPER.}
    \label{fig:gener}
\end{figure}

\begin{figure}[H]
    \centering
    \includegraphics[width=0.9\linewidth]{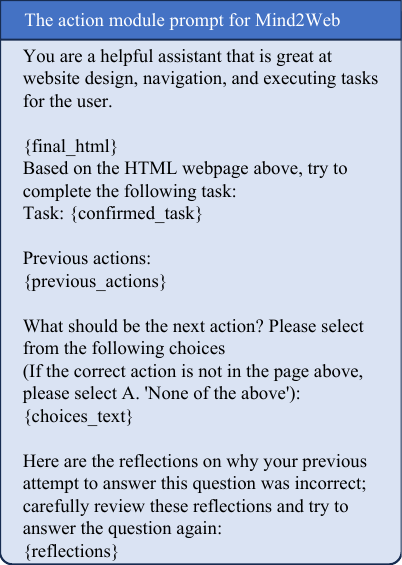}
    \caption{The action module prompt for Mind2Web when deploying Reflexion, Retroformer, and COPPER.}
    \label{fig:gener}
\end{figure}

\begin{figure*}[]
    \centering
    \includegraphics[width=0.9\linewidth]{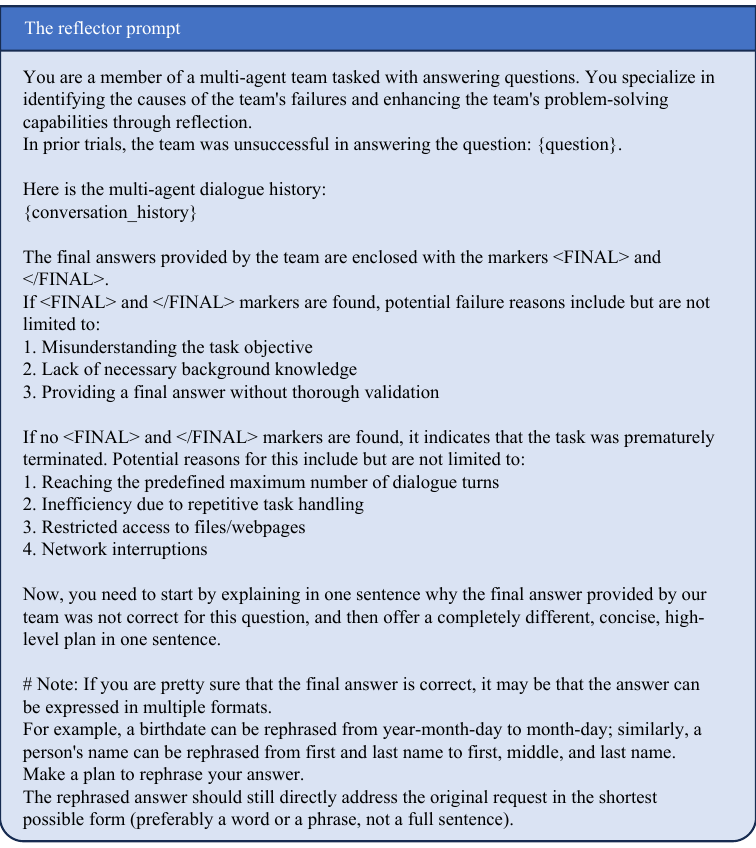}
    \caption{The reflector prompt when deploying Reflexion, Retroformer, and COPPER.}
    \label{fig:gener}
\end{figure*}

%% file: custom.bib
@inproceedings{yang2018hotpotqa,
  title={HotpotQA: A Dataset for Diverse, Explainable Multi-hop Question Answering},
  author={Yang, Zhilin and Qi, Peng and Zhang, Saizheng and Bengio, Yoshua and Cohen, William and Salakhutdinov, Ruslan and Manning, Christopher D},
  booktitle={Proceedings of the 2018 Conference on Empirical Methods in Natural Language Processing},
  pages={2369--2380},
  year={2018}
}

@inproceedings{qin-etal-2019-counterfactual,
    title = "Counterfactual Story Reasoning and Generation",
    author = "Qin, Lianhui  and
      Bosselut, Antoine  and
      Holtzman, Ari  and
      Bhagavatula, Chandra  and
      Clark, Elizabeth  and
      Choi, Yejin",
    editor = "Inui, Kentaro  and
      Jiang, Jing  and
      Ng, Vincent  and
      Wan, Xiaojun",
    booktitle = "Proceedings of the 2019 Conference on Empirical Methods in Natural Language Processing and the 9th Internatyang2018hotpotqaional Joint Conference on Natural Language Processing (EMNLP-IJCNLP)",
    month = nov,
    year = "2019",
    address = "Hong Kong, China",
    publisher = "Association for Computational Linguistics",
    url = "https://aclanthology.org/D19-1509/",
    doi = "10.18653/v1/D19-1509",
    pages = "5043--5053"
}

@article{bottou2013counterfactual,
  title={Counterfactual reasoning and learning systems: The example of computational advertising},
  author={Bottou, L{\'e}on and Peters, Jonas and Qui{\~n}onero-Candela, Joaquin and Charles, Denis X and Chickering, D Max and Portugaly, Elon and Ray, Dipankar and Simard, Patrice and Snelson, Ed},
  journal={The Journal of Machine Learning Research},
  volume={14},
  number={1},
  pages={3207--3260},
  year={2013},
  publisher={JMLR. org}
}

@inproceedings{zhang-etal-2025-lessons,
    title = "The Lessons of Developing Process Reward Models in Mathematical Reasoning",
    author = "Zhang, Zhenru  and
      Zheng, Chujie  and
      Wu, Yangzhen  and
      Zhang, Beichen  and
      Lin, Runji  and
      Yu, Bowen  and
      Liu, Dayiheng  and
      Zhou, Jingren  and
      Lin, Junyang",
    editor = "Che, Wanxiang  and
      Nabende, Joyce  and
      Shutova, Ekaterina  and
      Pilehvar, Mohammad Taher",
    booktitle = "Findings of the Association for Computational Linguistics: ACL 2025",
    month = jul,
    year = "2025",
    address = "Vienna, Austria",
    publisher = "Association for Computational Linguistics",
    url = "https://aclanthology.org/2025.findings-acl.547/",
    pages = "10495--10516",
    ISBN = "979-8-89176-256-5"
}

@inproceedings{
yao2024retroformer,
title={Retroformer: Retrospective Large Language Agents with Policy Gradient Optimization},
author={Weiran Yao and Shelby Heinecke and Juan Carlos Niebles and Zhiwei Liu and Yihao Feng and Le Xue and Rithesh R N and Zeyuan Chen and Jianguo Zhang and Devansh Arpit and Ran Xu and Phil L Mui and Huan Wang and Caiming Xiong and Silvio Savarese},
booktitle={The Twelfth International Conference on Learning Representations},
year={2024},
url={https://openreview.net/forum?id=KOZu91CzbK}
}

@article{shinn2023reflexion,
  title={Reflexion: Language agents with verbal reinforcement learning},
  author={Shinn, Noah and Cassano, Federico and Gopinath, Ashwin and Narasimhan, Karthik and Yao, Shunyu},
  journal={Advances in Neural Information Processing Systems},
  volume={36},
  pages={8634--8652},
  year={2023}
}

@article{deng2023mind2web,
  title={Mind2web: Towards a generalist agent for the web},
  author={Deng, Xiang and Gu, Yu and Zheng, Boyuan and Chen, Shijie and Stevens, Sam and Wang, Boshi and Sun, Huan and Su, Yu},
  journal={Advances in Neural Information Processing Systems},
  volume={36},
  pages={28091--28114},
  year={2023}
}

@inproceedings{wu2024autogen,
  title={AutoGen: Enabling Next-Gen LLM Applications via Multi-Agent Conversation},
  author={Wu, Qingyun and Bansal, Gagan and Zhang, Jieyu and Wu, Yiran and Li, Beibin and Zhu, Erkang and Jiang, Li and Zhang, Xiaoyun and Zhang, Shaokun and Liu, Jiale and others},
  booktitle={ICLR 2024 Workshop on Large Language Model (LLM) Agents},
year={2024}
}

@inproceedings{qian2024chatdev,
  title={Chatdev: Communicative agents for software development},
  author={Qian, Chen and Liu, Wei and Liu, Hongzhang and Chen, Nuo and Dang, Yufan and Li, Jiahao and Yang, Cheng and Chen, Weize and Su, Yusheng and Cong, Xin and others},
  booktitle={Proceedings of the 62nd Annual Meeting of the Association for Computational Linguistics (Volume 1: Long Papers)},
  pages={15174--15186},
  year={2024}
}

@inproceedings{chen2023agentverse,
  title={Agentverse: Facilitating multi-agent collaboration and exploring emergent behaviors},
  author={Chen, Weize and Su, Yusheng and Zuo, Jingwei and Yang, Cheng and Yuan, Chenfei and Chan, Chi-Min and Yu, Heyang and Lu, Yaxi and Hung, Yi-Hsin and Qian, Chen and others},
  booktitle={The Twelfth International Conference on Learning Representations},
  year={2023}
}

@inproceedings{hong2024metagpt,
      title={Meta{GPT}: Meta Programming for A Multi-Agent Collaborative Framework},
      author={Sirui Hong and Mingchen Zhuge and Jonathan Chen and Xiawu Zheng and Yuheng Cheng and Jinlin Wang and Ceyao Zhang and Zili Wang and Steven Ka Shing Yau and Zijuan Lin and Liyang Zhou and Chenyu Ran and Lingfeng Xiao and Chenglin Wu and J{\"u}rgen Schmidhuber},
      booktitle={The Twelfth International Conference on Learning Representations},
      year={2024},
      url={https://openreview.net/forum?id=VtmBAGCN7o}
}

@inproceedings{zhao2024expel,
  title={Expel: Llm agents are experiential learners},
  author={Zhao, Andrew and Huang, Daniel and Xu, Quentin and Lin, Matthieu and Liu, Yong-Jin and Huang, Gao},
  booktitle={Proceedings of the AAAI Conference on Artificial Intelligence},
  volume={38},
  number={17},
  pages={19632--19642},
  year={2024}
}

@article{fourney2024magentic,
  title={Magentic-one: A generalist multi-agent system for solving complex tasks},
  author={Fourney, Adam and Bansal, Gagan and Mozannar, Hussein and Tan, Cheng and Salinas, Eduardo and Niedtner, Friederike and Proebsting, Grace and Bassman, Griffin and Gerrits, Jack and Alber, Jacob and others},
  journal={arXiv preprint arXiv:2411.04468},
  year={2024}
}

@inproceedings{trivedi2024appworld,
  title={AppWorld: A Controllable World of Apps and People for Benchmarking Interactive Coding Agents},
  author={Trivedi, Harsh and Khot, Tushar and Hartmann, Mareike and Manku, Ruskin and Dong, Vinty and Li, Edward and Gupta, Shashank and Sabharwal, Ashish and Balasubramanian, Niranjan},
  booktitle={Proceedings of the 62nd Annual Meeting of the Association for Computational Linguistics (Volume 1: Long Papers)},
  pages={16022--16076},
  year={2024}
}

@inproceedings{lightman2023let,
  title={Let's verify step by step},
  author={Lightman, Hunter and Kosaraju, Vineet and Burda, Yuri and Edwards, Harrison and Baker, Bowen and Lee, Teddy and Leike, Jan and Schulman, John and Sutskever, Ilya and Cobbe, Karl},
  booktitle={The Twelfth International Conference on Learning Representations},
  year={2023}
}

@inproceedings{
hu2022lora,
title={Lo{RA}: Low-Rank Adaptation of Large Language Models},
author={Edward J Hu and Yelong Shen and Phillip Wallis and Zeyuan Allen-Zhu and Yuanzhi Li and Shean Wang and Lu Wang and Weizhu Chen},
booktitle={International Conference on Learning Representations},
year={2022},
url={https://openreview.net/forum?id=nZeVKeeFYf9}
}

@inproceedings{wang2024rethinking,
  title={Rethinking the Bounds of LLM Reasoning: Are Multi-Agent Discussions the Key?},
  author={Wang, Qineng and Wang, Zihao and Su, Ying and Tong, Hanghang and Song, Yangqiu},
  booktitle={Proceedings of the 62nd Annual Meeting of the Association for Computational Linguistics (Volume 1: Long Papers)},
  pages={6106--6131},
  year={2024}
}

@inproceedings{yang2024unveiling,
  title={Unveiling the Generalization Power of Fine-Tuned Large Language Models},
  author={Yang, Haoran and Zhang, Yumeng and Xu, Jiaqi and Lu, Hongyuan and Heng, Pheng-Ann and Lam, Wai},
  booktitle={Proceedings of the 2024 Conference of the North American Chapter of the Association for Computational Linguistics: Human Language Technologies (Volume 1: Long Papers)},
  pages={884--899},
  year={2024}
}

@inproceedings{
pan2025why,
title={Why Do Multiagent Systems Fail?},
author={Melissa Z Pan and Mert Cemri and Lakshya A Agrawal and Shuyi Yang and Bhavya Chopra and Rishabh Tiwari and Kurt Keutzer and Aditya Parameswaran and Kannan Ramchandran and Dan Klein and Joseph E. Gonzalez and Matei Zaharia and Ion Stoica},
booktitle={ICLR 2025 Workshop on Building Trust in Language Models and Applications},
year={2025},
url={https://openreview.net/forum?id=wM521FqPvI}
}

@article{zhang2025if,
  title={If multi-agent debate is the answer, what is the question},
  author={Zhang, Hangfan and Cui, Zhiyao and Wang, Xinrun and Zhang, Qiaosheng and Wang, Zhen and Wu, Dinghao and Hu, Shuyue},
  journal={arXiv preprint arXiv:2502.08788},
  year={2025}
}

@article{bo2024reflective,
  title={Reflective multi-agent collaboration based on large language models},
  author={Bo, Xiaohe and Zhang, Zeyu and Dai, Quanyu and Feng, Xueyang and Wang, Lei and Li, Rui and Chen, Xu and Wen, Ji-Rong},
  journal={Advances in Neural Information Processing Systems},
  volume={37},
  pages={138595--138631},
  year={2024}
}

@article{wang2024survey,
  title={A survey on large language model based autonomous agents},
  author={Wang, Lei and Ma, Chen and Feng, Xueyang and Zhang, Zeyu and Yang, Hao and Zhang, Jingsen and Chen, Zhiyuan and Tang, Jiakai and Chen, Xu and Lin, Yankai and others},
  journal={Frontiers of Computer Science},
  volume={18},
  number={6},
  pages={186345},
  year={2024},
  publisher={Springer}
}

@article{schulman2017proximal,
  title={Proximal policy optimization algorithms},
  author={Schulman, John and Wolski, Filip and Dhariwal, Prafulla and Radford, Alec and Klimov, Oleg},
  journal={arXiv preprint arXiv:1707.06347},
  year={2017}
}

@article{ouyang2022training,
  title={Training language models to follow instructions with human feedback},
  author={Ouyang, Long and Wu, Jeffrey and Jiang, Xu and Almeida, Diogo and Wainwright, Carroll and Mishkin, Pamela and Zhang, Chong and Agarwal, Sandhini and Slama, Katarina and Ray, Alex and others},
  journal={Advances in neural information processing systems},
  volume={35},
  pages={27730--27744},
  year={2022}
}

@inproceedings{
zhang2025which,
title={Which Agent Causes Task Failures and When? On Automated Failure Attribution of {LLM} Multi-Agent Systems},
author={Shaokun Zhang and Ming Yin and Jieyu Zhang and Jiale Liu and Zhiguang Han and Jingyang Zhang and Beibin Li and Chi Wang and Huazheng Wang and Yiran Chen and Qingyun Wu},
booktitle={Forty-second International Conference on Machine Learning},
year={2025},
url={https://openreview.net/forum?id=GazlTYxZss}
}

@inproceedings{masry2025chartqapro,
    title = "{C}hart{QAP}ro: A More Diverse and Challenging Benchmark for Chart Question Answering",
    author = "Masry, Ahmed  and
      Islam, Mohammed Saidul  and
      Ahmed, Mahir  and
      Bajaj, Aayush  and
      Kabir, Firoz  and
      Kartha, Aaryaman  and
      Laskar, Md Tahmid Rahman  and
      Rahman, Mizanur  and
      Rahman, Shadikur  and
      Shahmohammadi, Mehrad  and
      Thakkar, Megh  and
      Parvez, Md Rizwan  and
      Hoque, Enamul  and
      Joty, Shafiq",
    editor = "Che, Wanxiang  and
      Nabende, Joyce  and
      Shutova, Ekaterina  and
      Pilehvar, Mohammad Taher",
    booktitle = "Findings of the Association for Computational Linguistics: ACL 2025",
    month = jul,
    year = "2025",
    address = "Vienna, Austria",
    publisher = "Association for Computational Linguistics",
    url = "https://aclanthology.org/2025.findings-acl.978/",
    doi = "10.18653/v1/2025.findings-acl.978",
    pages = "19123--19151",
    ISBN = "979-8-89176-256-5"
}

@inproceedings{song-etal-2025-prmbench,
    title = "{PRMB}ench: A Fine-grained and Challenging Benchmark for Process-Level Reward Models",
    author = "Song, Mingyang  and
      Su, Zhaochen  and
      Qu, Xiaoye  and
      Zhou, Jiawei  and
      Cheng, Yu",
    editor = "Che, Wanxiang  and
      Nabende, Joyce  and
      Shutova, Ekaterina  and
      Pilehvar, Mohammad Taher",
    booktitle = "Proceedings of the 63rd Annual Meeting of the Association for Computational Linguistics (Volume 1: Long Papers)",
    month = jul,
    year = "2025",
    address = "Vienna, Austria",
    publisher = "Association for Computational Linguistics",
    url = "https://aclanthology.org/2025.acl-long.1230/",
    pages = "25299--25346",
    ISBN = "979-8-89176-251-0"
}

@inproceedings{
wang2025openhands,
title={OpenHands: An Open Platform for {AI} Software Developers as Generalist Agents},
author={Xingyao Wang and Boxuan Li and Yufan Song and Frank F. Xu and Xiangru Tang and Mingchen Zhuge and Jiayi Pan and Yueqi Song and Bowen Li and Jaskirat Singh and Hoang H. Tran and Fuqiang Li and Ren Ma and Mingzhang Zheng and Bill Qian and Yanjun Shao and Niklas Muennighoff and Yizhe Zhang and Binyuan Hui and Junyang Lin and Robert Brennan and Hao Peng and Heng Ji and Graham Neubig},
booktitle={The Thirteenth International Conference on Learning Representations},
year={2025},
url={https://openreview.net/forum?id=OJd3ayDDoF}
}

@article{swanson2024virtual,
  title={The virtual lab: Ai agents design new sars-cov-2 nanobodies with experimental validation},
  author={Swanson, Kyle and Wu, Wesley and Bulaong, Nash L and Pak, John E and Zou, James},
  journal={bioRxiv},
  pages={2024--11},
  year={2024},
  publisher={Cold Spring Harbor Laboratory}
}

@article{gottweis2025towards,
  title={Towards an AI co-scientist},
  author={Gottweis, Juraj and Weng, Wei-Hung and Daryin, Alexander and Tu, Tao and Palepu, Anil and Sirkovic, Petar and Myaskovsky, Artiom and Weissenberger, Felix and Rong, Keran and Tanno, Ryutaro and others},
  journal={arXiv preprint arXiv:2502.18864},
  year={2025}
}

@article{li2023camel,
  title={Camel: Communicative agents for" mind" exploration of large language model society},
  author={Li, Guohao and Hammoud, Hasan and Itani, Hani and Khizbullin, Dmitrii and Ghanem, Bernard},
  journal={Advances in Neural Information Processing Systems},
  volume={36},
  pages={51991--52008},
  year={2023}
}

@inproceedings{zhang2024self,
  title={Self-Contrast: Better Reflection Through Inconsistent Solving Perspectives},
  author={Zhang, Wenqi and Shen, Yongliang and Wu, Linjuan and Peng, Qiuying and Wang, Jun and Zhuang, Yueting and Lu, Weiming},
  booktitle={Proceedings of the 62nd Annual Meeting of the Association for Computational Linguistics (Volume 1: Long Papers)},
  pages={3602--3622},
  year={2024}
}

@inproceedings{
du2024improving,
title={Improving Factuality and Reasoning in Language Models through Multiagent Debate},
author={Yilun Du and Shuang Li and Antonio Torralba and Joshua B. Tenenbaum and Igor Mordatch},
booktitle={Forty-first International Conference on Machine Learning},
year={2024},
url={https://openreview.net/forum?id=zj7YuTE4t8}
}

@inproceedings{
chen2024teaching,
title={Teaching Large Language Models to Self-Debug},
author={Xinyun Chen and Maxwell Lin and Nathanael Sch{\"a}rli and Denny Zhou},
booktitle={The Twelfth International Conference on Learning Representations},
year={2024},
url={https://openreview.net/forum?id=KuPixIqPiq}
}

@article{madaan2023self,
  title={Self-refine: Iterative refinement with self-feedback},
  author={Madaan, Aman and Tandon, Niket and Gupta, Prakhar and Hallinan, Skyler and Gao, Luyu and Wiegreffe, Sarah and Alon, Uri and Dziri, Nouha and Prabhumoye, Shrimai and Yang, Yiming and others},
  journal={Advances in Neural Information Processing Systems},
  volume={36},
  pages={46534--46594},
  year={2023}
}

@article{ghafarollahi2025sciagents,
  title={SciAgents: automating scientific discovery through bioinspired multi-agent intelligent graph reasoning},
  author={Ghafarollahi, Alireza and Buehler, Markus J},
  journal={Advanced Materials},
  volume={37},
  number={22},
  pages={2413523},
  year={2025},
  publisher={Wiley Online Library}
}

@article{tang2023medagents,
  title={Medagents: Large language models as collaborators for zero-shot medical reasoning},
  author={Tang, Xiangru and Zou, Anni and Zhang, Zhuosheng and Li, Ziming and Zhao, Yilun and Zhang, Xingyao and Cohan, Arman and Gerstein, Mark},
  journal={arXiv preprint arXiv:2311.10537},
  year={2023}
}
